\documentclass[10pt,twocolumn,letterpaper]{article}

\usepackage{booktabs}
\usepackage{multirow}
\usepackage{stfloats}
\usepackage{soul}
\usepackage[ruled,linesnumbered]{algorithm2e}
\usepackage[pagenumbers]{cvpr} % To force page numbers, e.g. for an arXiv version

\definecolor{cvprblue}{rgb}{0.21,0.49,0.74}
\usepackage[pagebackref,breaklinks,colorlinks,allcolors=cvprblue]{hyperref}

\def\paperID{5629} % *** Enter the Paper ID here
\def\confName{CVPR}
\def\confYear{2025}

\title{TSD-SR: One-Step Diffusion with Target Score Distillation for Real-World Image Super-Resolution}

\def\spaces{~~~~~~}
\author{
Linwei Dong\textsuperscript{1,2}\thanks{Equal contribution. ~~\textsuperscript{\dag}Corresponding authors.}\spaces{}
Qingnan Fan\textsuperscript{2}\footnotemark[1]\spaces{}
Yihong Guo\textsuperscript{1}\spaces{}
Zhonghao Wang\textsuperscript{3}\spaces{} \\
Qi Zhang\textsuperscript{2}\spaces{}
Jinwei Chen\textsuperscript{2}\spaces{}
Yawei Luo\textsuperscript{1}$^{\dagger}$\spaces{}
Changqing Zou\textsuperscript{1,4}\\\\
\textsuperscript{1}Zhejiang University\spaces{}
\textsuperscript{2}Vivo Mobile Communication Co. Ltd\spaces{} \\
\textsuperscript{3}University of Chinese Academy of Sciences \spaces{}
\textsuperscript{4}Zhejiang Lab
}

\begin{document}
\maketitle
\begin{abstract}
Pre-trained text-to-image diffusion models are increasingly applied to real-world image super-resolution (Real-ISR) tasks. Given the iterative refinement nature of diffusion models, most existing approaches are computationally expensive. While methods such as SinSR and OSEDiff have emerged to condense inference steps via distillation, their performance in image restoration or details recovery is not satisfactory. To address this, we propose TSD-SR, a novel distillation framework specifically designed for real-world image super-resolution, aiming to construct an efficient and effective one-step model. 
We first introduce the Target Score Distillation, which leverages the priors of diffusion models and real image references to achieve more realistic image restoration. Secondly, we propose a Distribution-Aware Sampling Module to make detail-oriented gradients more readily accessible, addressing the challenge of recovering fine details. Extensive experiments demonstrate that our TSD-SR has superior restoration results (most of the metrics perform the best) and the fastest inference speed (e.g. 40 times faster than SeeSR) compared to the past Real-ISR approaches based on pre-trained diffusion priors. Our code is released at \href{https://github.com/Microtreei/TSD-SR}{\textcolor{pink}{https://github.com/Microtreei/TSD-SR}}.
\end{abstract}    
\section{Introduction}
\label{sec:intro}

% Image super-resolution (ISR) \cite{dong2015image,dong2014learning,kim2016accurate,ledig2017photo} aims to transform low-quality (LQ) images, which have experienced noise or blur, into clear high-quality (HQ) images. Differing from traditional ISR \cite{chen2021pre,zhang2022efficient}, which assumes a known degradation process, real-world image super-resolution (Real-ISR) \cite{wang2021real,zhang2021designing} is designed to enhance real-world images that have suffered from complex and unknown degradations, thereby offering greater practical utility.
Image super-resolution (ISR) \cite{dong2015image,dong2014learning,kim2016accurate,ledig2017photo} aims to transform low-quality (LQ) images, which have been degraded by noise or blur, into clear high-quality (HQ) images. Unlike traditional ISR methods \cite{chen2021pre,zhang2022efficient}, which assume a known degradation process, real-world image super-resolution (Real-ISR) \cite{wang2021real,zhang2021designing} focuses on enhancing images affected by complex and unknown degradations, thereby offering greater practical utility.

%-------------------------------------------------------------------------
% Generative models, particularly Generative Adversarial Networks (GANs) \cite{goodfellow2014generative, mirza2014conditional, radford2015unsupervised} and Diffusion Models (DMs) \cite{song2020score,ho2020denoising,rombach2022high}, have shown remarkable power in handling Real-ISR tasks. GAN-based methods leverage adversarial training, toggling between the generator and discriminator to produce realistic images. While GANs are capable of one-step inference, they are commonly hampered by issues like mode collapse and training instability \cite{arjovsky2017wasserstein}.
Generative models, particularly Generative Adversarial Networks (GANs) \cite{goodfellow2014generative, mirza2014conditional, radford2015unsupervised} and Diffusion Models (DMs) \cite{song2020score,ho2020denoising,rombach2022high}, have demonstrated remarkable capabilities in tackling Real-ISR tasks. GAN-based methods utilize adversarial training by alternately optimizing a generator and a discriminator to produce realistic images. While GANs support one-step inference, they are often hindered by challenges such as mode collapse and training instability \cite{arjovsky2017wasserstein}.
%-------------------------------------------------------------------------
% Recently, Diffusion Models (DMs) have shown impressive performance in the realm of image generation \cite{wang2022zero, kawar2022denoising}. Their robust priors empower them to produce more realistic images with richer details than GAN-based methods \cite{rombach2022high, song2020score}. Some researchers \cite{lin2023diffbir, yu2024scaling, wu2024seesr,yang2023pixel} have successfully leveraged pre-trained DMs for Real-ISR tasks. However, due to the iterative denoising nature of diffusion models \cite{ho2020denoising}, the Real-ISR process is computationally expensive.
Recently, Diffusion Models (DMs) have demonstrated impressive performance in image generation \cite{wang2022zero, kawar2022denoising}. Their strong priors enable them to produce more realistic images with richer details compared to GAN-based methods \cite{rombach2022high, song2020score}. Some researchers \cite{lin2023diffbir, yu2024scaling, wu2024seesr, yang2023pixel} have successfully leveraged pre-trained DMs for Real-ISR tasks. However, due to the iterative denoising nature of diffusion models \cite{ho2020denoising}, the Real-ISR process is computationally expensive.

%-------------------------------------------------------------------------
%To achieve an efficient and one-step network akin to GANs, several pioneering methods that condense the iterations of diffusion models through distillation have been proposed\cite{wu2024one,xie2024addsr}. Among these works, OSEDiff introduces a Variational Score Distillation (VSD) loss initiatively to Real-ISR tasks and achieves state-of-the-art (SOTA) one-step results. This approach leverages the pre-trained model's prior knowledge to optimize the learning process. However, it may encounter the following two obstacles: {\bf (1) Unreliable gradient direction}.  VSD uses a Teacher Model to give a “true gradient direction”. Nevertheless, this direction is unreliable when encountering poor ISR results. {\bf (2) Detail recovery is insufficient}. In Real-ISR, there is divergence in the VSD loss across different timesteps. The uniform sampling strategy for $t$ results in difficulty matching the scores function that focuses on detail recovery.

\begin{figure}[!t]
  \centering
  \setlength{\abovecaptionskip}{0.1cm}
  \setlength{\belowcaptionskip}{-0.5cm}
    \includegraphics[width=\linewidth]{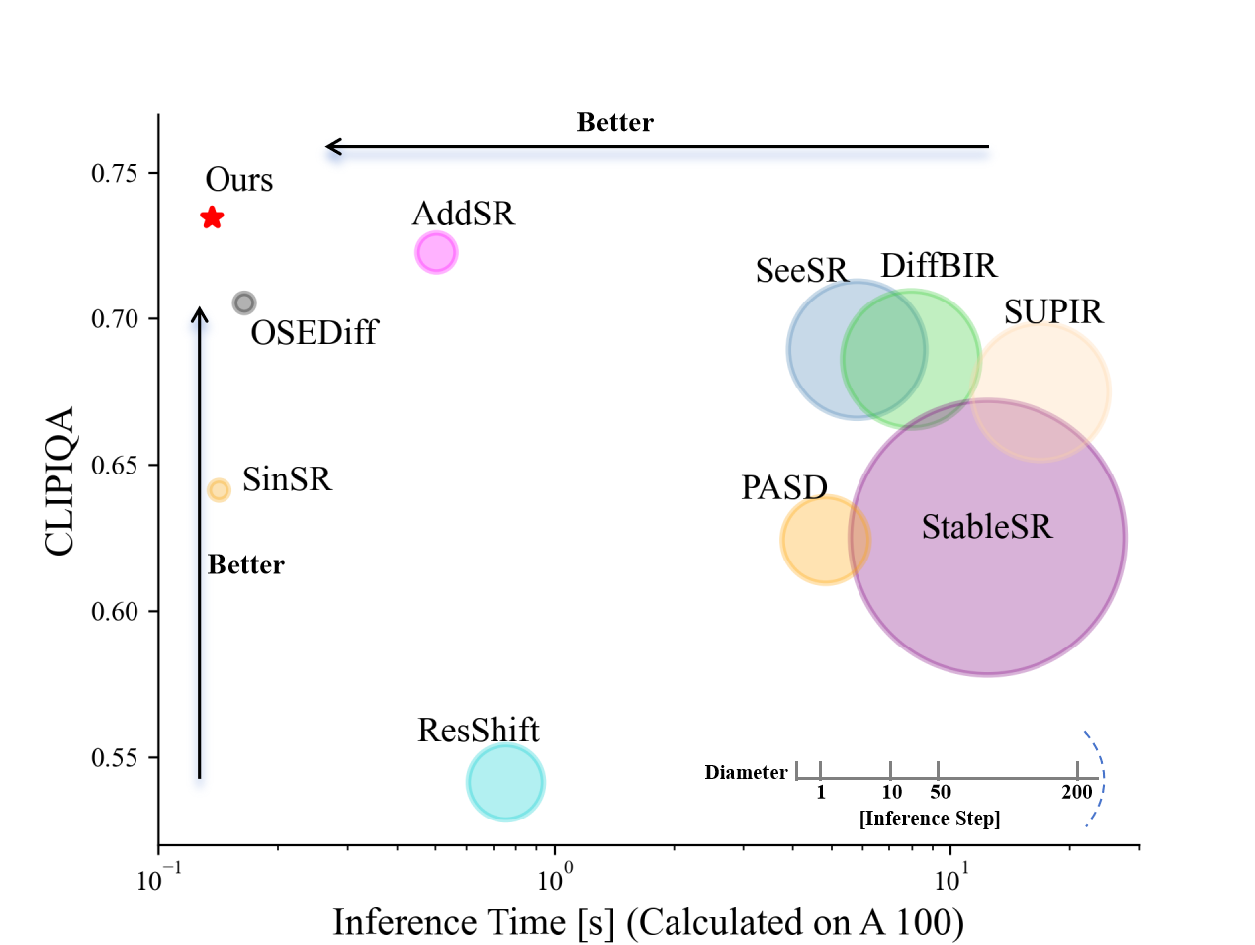}
   \caption{Performance and efficiency comparison among Real-ISR methods. TSD-SR stands out for achieving high-quality restoration with the fastest speed among diffusion-based models. In contrast, existing models prioritize either speed or restoration performance. The performance of each method is benchmarked on an A100 GPU with the DRealSR dataset. }
   \label{fig:bubble}
\end{figure}

To achieve an efficient and one-step network akin to GANs, several pioneering methods that condense the iterations of diffusion models through distillation \cite{hinton2015distilling,yim2017gift,gou2021knowledge,howard2017mobilenets} have been proposed \cite{wu2024one,xie2024addsr, wang2024sinsr}. 
Among these works, OSEDiff \cite{wu2024one} introduced the Variational Score Distillation (VSD) loss \cite{wang2024prolificdreamer} to Real-ISR tasks, achieving state-of-the-art (SOTA) one-step performance by leveraging prior knowledge from pre-trained models. Despite these advancements, our investigation has revealed two critical limitations associated with VSD in Real-ISR applications. {\bf (1) Unreliable gradient direction.} VSD relies on a Teacher Model to provide a ``true gradient direction.'' However, this guidance is proven unreliable in scenarios where initial ISR outputs are suboptimal. {\bf (2) Insufficient detail recovery.} The VSD loss exhibits notable variation across different timesteps, and the uniform sampling strategy for $t$ poses challenges in aligning the score function with detailed texture recovery requirements. These findings underscore the need for more effective approaches to address these issues.

%-------------------------------------------------------------------------

In this paper, we propose a novel method called  {\bf TSD-SR} to distill a multi-step Text-to-Image (T2I) DMs \cite{ramesh2022hierarchical, rombach2022high, esser2024scaling} into an effective one-step diffusion model tailored for the Real-ISR task.
Specifically, TSD-SR consists of two components: {\bf Target Score Distillation (TSD)} and {\bf Distribution-Aware Sampling Module (DASM)}.
TSD incorporates our newly proposed Target Score Matching (TSM) loss to compensate for the limitations of the VSD loss. This significant score loss leverages HQ data to provide a reliable optimization trajectory during distillation, effectively reducing visual artifacts caused by deviant predictions from the Teacher Model.
DASM is designed to enhance detail recovery by strategically sampling low-noise samples that are distribution-based during training. This approach effectively allocates more optimization to early timesteps within a single iteration, thereby improving the recovery of fine details.

Experiments on popular benchmarks demonstrate that TSD-SR achieves superior restoration performance (most of the metrics perform the best) and high efficiency (the fastest inference speed, 40 times faster than SeeSR) compared to the state-of-the-art Real-ISR methods based on pre-trained DMs, while requiring only a single inference step, as shown in (\cref{fig:bubble}). 
% Experiments on popular benchmarks demonstrate that TSD-SR achieves superior restoration performance, achieving the best results on most metrics, and high efficiency, with the fastest inference speed—40 times faster than SeeSR—compared to state-of-the-art Real-ISR methods based on pre-trained diffusion models, while requiring only a single inference step (\cref{fig:bubble}).
%-------------------------------------------------------------------------

\noindent
Our main contribution can be summarized as threefold:
\begin{itemize}
\item We propose a novel method called TSD-SR  to achieve one-step DMs distillation for the Real-ISR task.
\item We introduce Target Score Distillation (TSD) to provide reliable gradients that enhance the realism of outputs from Real-ISR methods.
\item We design a Distribution-Aware Sampling Module (DASM) specifically tailored to enhance the capability of detail restoration.
\end{itemize}

\section{Related Work}
\label{sec:related}
{\bf GAN-based Real-ISR.}
Since SRGAN \cite{ledig2017photo} first applied GAN to ISR, it has effectively enhanced visual quality by combining adversarial loss with perceptual loss \cite{zhang2018unreasonable, ding2020image}. Subsequently, ESRGAN \cite{wang2018esrgan} introduced Residual-in-Residual Dense Block and a relativistic average discriminator, further improving detail restoration. Methods like BSRGAN \cite{zhang2021designing} and Real-ESRGAN \cite{wang2021real} simulate complex real-world degradation processes, achieving ISR under unknown degradation conditions, which enhances the model's generalization ability. 
Although GAN-based methods are capable of adding more realistic details to images, they suffer from training instability and mode collapse \cite{arjovsky2017wasserstein}.

\noindent
{\bf Multi-step Diffusion-based Real-ISR.}
Some researches \cite{wang2024exploiting,lin2023diffbir,wu2024seesr,yu2024scaling, yang2023pixel} in recent years have utilized the powerful image priors in pre-trained T2I diffusion models \cite{zhang2023adding,podell2023sdxl,rombach2022high} for Real-SR tasks and achieved promising results. 
For example, StableSR \cite{wang2024exploiting} balances fidelity and perceptual quality by fine-tuning the time-aware encoder and employing controllable feature wrapping. DiffBiR \cite{lin2023diffbir} first processes the LR image through a reconstruction network and then uses the Stable Diffusion (SD) model \cite{rombach2022high} to supplement the details. SeeSR \cite{wu2024seesr} attempts to better stimulate the generative power of the SD model by extracting the semantic information in the image as a conditional guide. PASD \cite{yang2023pixel} introduces a pixel-aware cross attention module to enable the diffusion model to perceive the local structure of the image at the pixel level, while using a degradation removal module to extract degradation insensitive features to guide the diffusion process along with high-level information from the image. SUPIR \cite{yu2024scaling} achieves a generative and fidelity capability using negative cues \cite{ho2022classifier} as well as restoration-guided sampling, while using a larger pre-training model with a larger dataset to enhance the model capability. 
However, all of these methods are limited by the multi-step denoising of the diffusion model, which requires 20-50 iterations in inference, resulting in an inference time that lags far behind that of GAN-based methods.

\noindent
{\bf One-step Diffusion-based Real-ISR.}
Recently, there has been a surge of interest within the academic community in one-step distillation techniques \cite{nguyen2024swiftbrush, yin2024one, sauer2023adversarial,yin2024improved,luo2023latent} for diffusion-based Real-ISR task.
SinSR \cite{wang2024sinsr} leverages consistency preserving distillation to condense the inference steps of ResShift \cite{yue2024resshift} into a single step, yet the generalization of ResShift and SinSR  is constrained due to the absence of large-scale data training. AddSR \cite{xie2024addsr} introduces the adversarial diffusion distillation (ADD) \cite{sauer2023adversarial} to Real-ISR tasks, resulting in a comparatively effective four-step model. However, this method has a propensity to produce excessive and unnatural image details. OSEDiff \cite{wu2024one} directly uses LQ images as the beginning of the diffusion process, and employs VSD loss \cite{wang2024prolificdreamer} as a regularization technique to condense a multi-step pre-trained T2I model into a one-step Real-ISR model. However, due to the incorporation of alternating training strategies, OSEDiff may initially tend towards unreliable optimization directions, which may lead to visual artifacts.

% \noindent
% {\bf Rectify Flow.}

\section{Methodology}
\label{sec:method}
% %-------------------------------------------------------------------------
\begin{figure*}[!htbp]
  \centering
    \includegraphics[width=\linewidth]{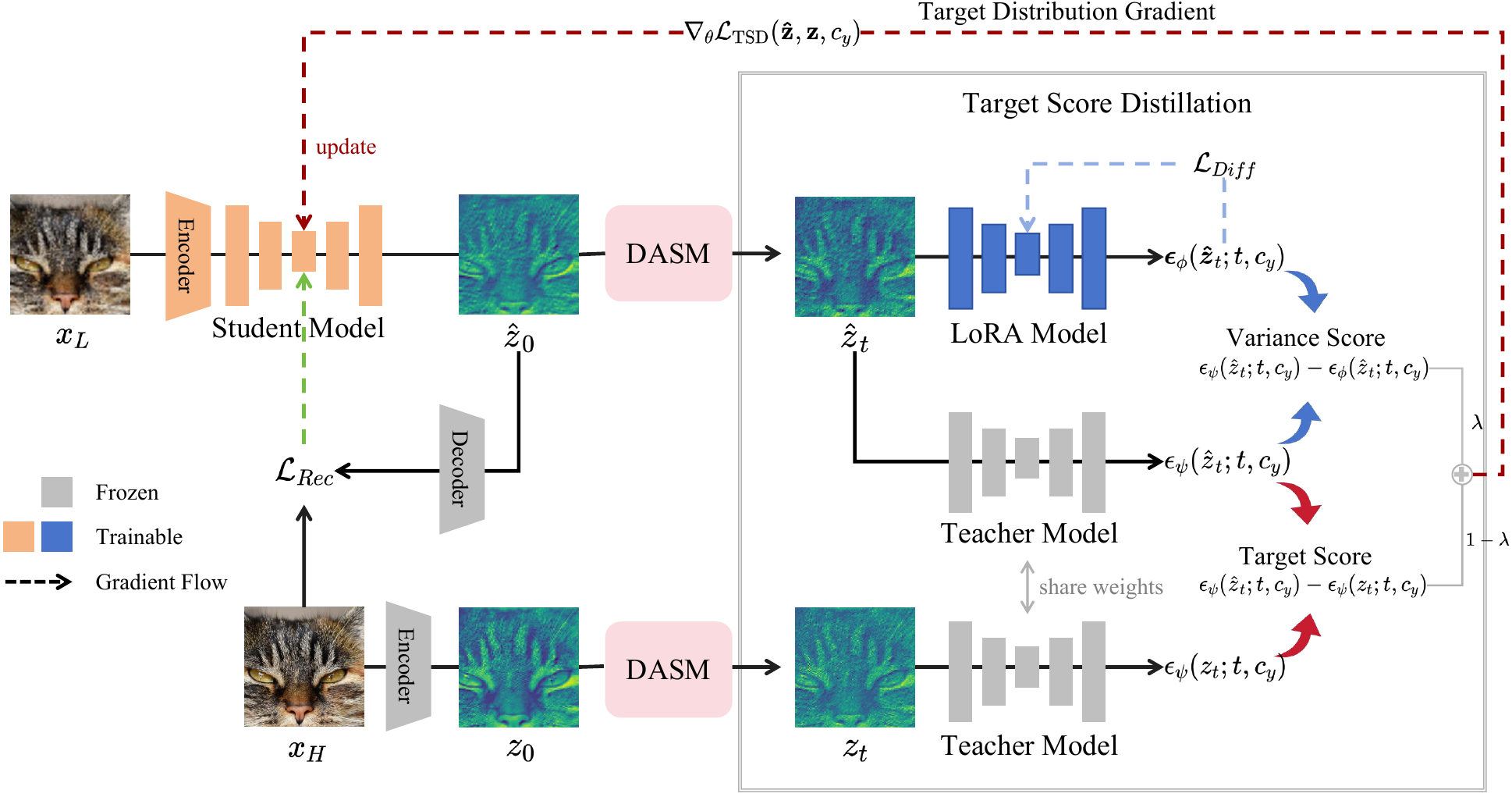}
   \caption{
Pipeline overview. We train a one-step Student Model $G_\theta$ to transform the low-quality image $x_L$ into a more realistic one. The noisy latent $\boldsymbol{\hat z_t}$ sampled by DASM (Details can be found in \cref{fig:DAMS}.) will be fed into both the pre-trained Teacher and the LoRA Model to produce the Variational Score Loss. Subsequently, the Teacher’s predictions on $\boldsymbol{\hat z_t}$ and $\boldsymbol{z_t}$ yield the Target Score Loss. Their weighted forms, namely TSD ({\color[HTML]{d04e4e} {\textbf{red flow}}}), along with the pixel-space reconstruction loss ({\color[HTML]{6ab53a} {\textbf{green flow}}}), are leveraged to update the Student Model $G_\theta$ . After updating the Student Model, we employ the diffusion loss ({\color[HTML]{3f69c4} {\textbf{blue flow}}}) to update the LoRA Model. 
   }
   \label{fig:pipeline}
\end{figure*}

\subsection{Preliminaries} 
{\bf Problem Formulation.} The ISR problem aims to reconstruct a HQ image ${x}_H$ from an LQ input $x_L$  by training a parameterized ISR model $G_{\theta}$ on a dataset $\mathcal{D} = \{(x_L, x_H)_{i=1}^N\}$, where $N$  represents the number of image pairs. Formally, this problem can be formulated as minimizing the following objective:
\begin{equation}
\begin{aligned}
 \theta^*=\arg\min_\theta\mathbb{E}_{(x_L,x_H)\sim\mathcal{D}} [\mathcal{L}_{Rec}(G_\theta(x_L),x_H) \\
 +\lambda\mathcal{L}_{Reg}(q_{\theta}(\hat{x}_H),p(x_H))]
\end{aligned}
\label{eq:hr_object}
\end{equation}
\noindent
Here, $\mathcal{L}_{Rec}$ denotes the reconstruction loss, commonly measured by distance metrics such as $L_2$ or $LPIPS$ \cite{zhang2018unreasonable}. The regularization term $\mathcal{L}_{Reg}$ improves the realism and generalization of the output of the ISR model. This objective can be understood as aligning the ISR output $\hat x_H$'s distribution, $q_{\theta}(\hat{x}_H)$, with the high-quality data $x_H$'s distribution $p(x_H)$ by minimizing the KL-divergence \cite{kullback1951information}:
\begin{equation}
\min_{\theta} \mathcal{D}_{\mathrm{KL}}\left(q_\theta(\hat{x}_H) \| p(x_H)\right)
\label{eq:kl_object}
\end{equation}
While several studies \cite{wang2021real, zhang2021designing, wang2018esrgan} have employed adversarial loss to optimize this objective, they often encounter issues like mode collapse and training instability. Recent work \cite{wu2024one} achieved state-of-the-art results using Variational Score Distillation (VSD) as the regularization loss to minimize this objective, which inspires our research.

%-------------------------------------------------------------------------
%  Variational Score Distillation
\noindent
{\bf Variational Score Distillation.} 
Variational Score Distillation (VSD) \cite{wang2024prolificdreamer} was initially introduced for text-to-3D generation, by distilling a pre-trained text-to-image diffusion model to optimize a single 3D representation \cite{poole2022dreamfusion}. 

In the VSD framework, a pre-trained diffusion model, represented as $\boldsymbol{\epsilon}_{\psi}$, and its trainable (LoRA \cite{hu2021lora}) replica  $\boldsymbol{\epsilon}_{\phi}$, are used to regularize the generator network $G_{\theta}$. As outlined in ProlificDreamer \cite{wang2024prolificdreamer}, the gradient with respect to the generator parameters $\boldsymbol{\theta}$ is formulated as follows:
\begin{equation}
\begin{split}
& \nabla_{\boldsymbol{\theta}}\mathcal{L}_{\mathrm{VSD}}\left(\boldsymbol{\hat z},c_{y}\right) \\ 
&  = \mathbb{E}_{t,\epsilon}\left[\omega(t)\left(\boldsymbol{\epsilon}_{\psi}(\boldsymbol{\hat z_{t}};t,c_{y})-\boldsymbol{\epsilon}_{\phi}(\boldsymbol{\hat z_{t}};t,c_{y})\right)\frac{\partial\boldsymbol{\hat z}}{\partial\boldsymbol{\theta}}\right]
\end{split}
\label{eq:vsd}
\end{equation}

\noindent
where $\boldsymbol{\hat z_t} = \alpha_t \boldsymbol{\hat z} + \sigma_t \boldsymbol{\epsilon}$  is the noisy input, $\boldsymbol{\hat z}$ is the latent outputted by the generator network $G_{\theta}$, $\boldsymbol{\epsilon}$ is a Gaussian noise, and $\alpha_t, \sigma_t$  are the noise-data scaling constants. $c_y$ is a text embedding corresponding to a caption that describes the input image, and $w(t)$ is a time-varying weighting function.

%-------------------------------------------------------------------------
% Network Overview
\subsection{Overview of TSD-SR} 
As depicted in \cref{fig:pipeline},  our goal is to distill a given pre-trained T2I DM into a fast one-step Student Model $G_\theta$, using the Teacher Model $\boldsymbol{\epsilon}_\psi$ and the trainable LoRA Model $\boldsymbol{\epsilon}_\phi$.  We denote the latent output of the distilled model as $\boldsymbol{\hat z_0}$, and the HQ latent representation as $\boldsymbol{ z_0}$. Both  $\boldsymbol{\hat z_0}$  and  $\boldsymbol{ z_0}$ are passed through our Distribution-Aware Sampling Module (DASM) to obtain distribution-based samples $\boldsymbol{\hat z_t}$ and $\boldsymbol{ z_t}$ (\cref{subsec:DASM}). We train $G_\theta$ by minimizing the two losses:  a reconstruction loss in pixel space to compare the model outputs against the ground truth, and a regularization loss (from Target Score Distillation) to enhance the realism (\cref{subsec:TSMD}). After updating the Student Model, we update the LoRA Model with the diffusion loss. Finally, in \cref{subsec:overall}, we present an overview of all the losses encountered during the training phase. 

%-------------------------------------------------------------------------
% Target Score Matching Distillation
\subsection{Target Score Distillation} 
\label{subsec:TSMD}
\begin{figure}[!t]
  \centering
    \includegraphics[width=\linewidth]{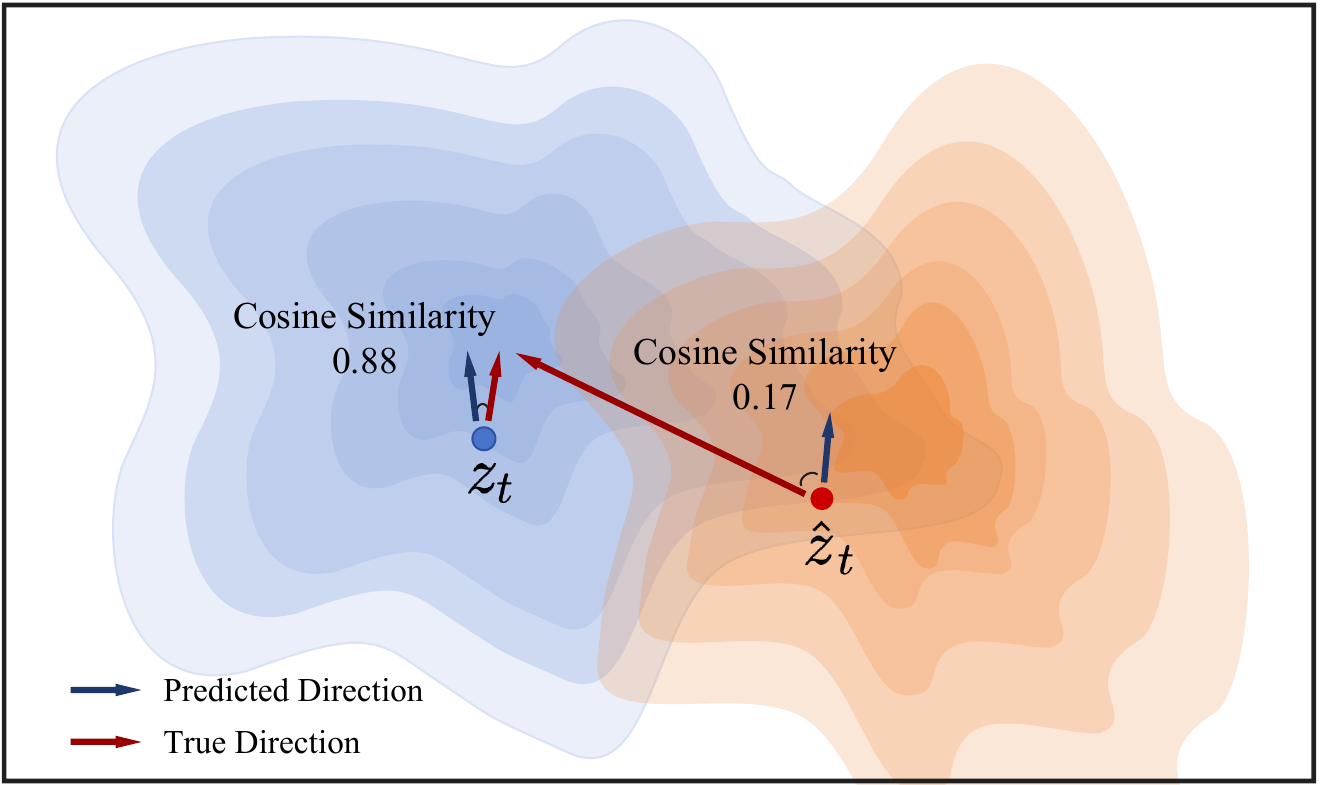}
   \caption{A visual comparison of the gradient direction. We set the timestep  $t$  to 100 and calculated the cosine similarity between the prediction directions from the Teacher Model and the true direction (towards the HQ data). The {\color[HTML]{3f69c4} {\textbf{prediction direction}}} for $\boldsymbol{z_t}$ closely matches the {\color[HTML]{c8182b} {\textbf{true direction}}}, but not for $\boldsymbol{\hat z_t}$, suggesting that suboptimal samples may lead to directional deviations.
}
\label{fig:gradient-figure}
\end{figure}

\begin{figure}[!t]
  \centering
  \begin{subfigure}{0.32\linewidth}
    \includegraphics[width=\linewidth]{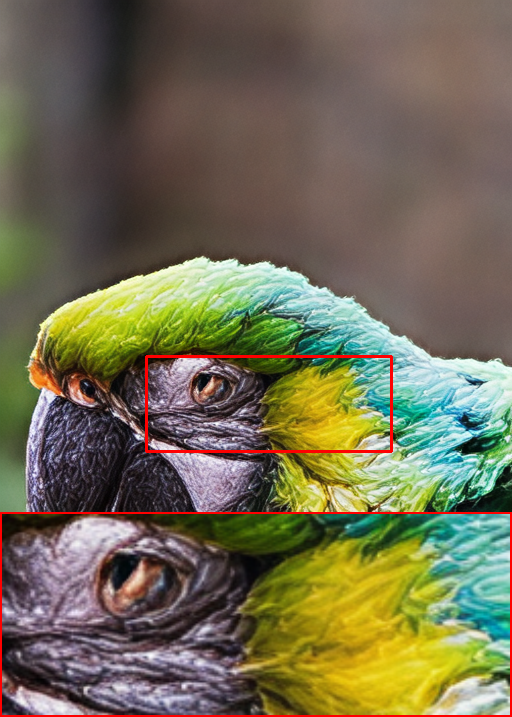}
    \caption{Naive}
    \label{fig:fig_vsd}
  \end{subfigure}
  \hfill
  \begin{subfigure}{0.32\linewidth }
    \includegraphics[width=\linewidth]{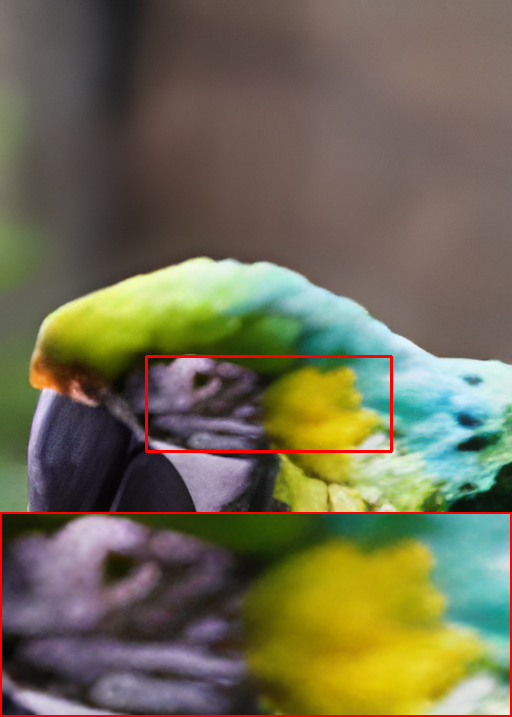}
    \caption{MSE}
    \label{fig:fig_mse}
  \end{subfigure}
\hfill
    \begin{subfigure}{0.32\linewidth }
    \includegraphics[width=\linewidth]{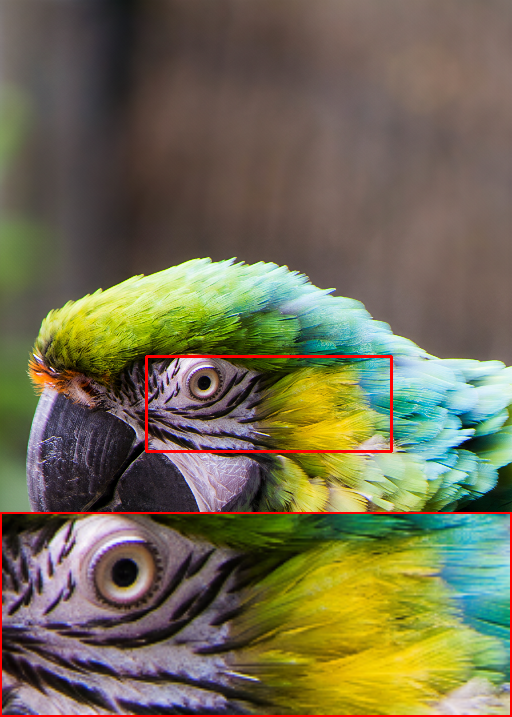}
    \caption{Ours}
    \label{fig:fig_ours}
  \end{subfigure}
  \caption{The visualization of different strategies. (a) The naive method introduces fake textures and fails to recover fine details.
(b) MSE leads to over-smoothed generation results, lacking high-frequency information.
(c) Our method offers the superior visual effects and fine textures.}
  % \vspace{-0.1cm}
  \label{fig:fig_visual}
\end{figure}

% {\bf Motivation.}
Similar to \cite{wu2024one}, we introduce VSD loss into our work as a regularization term to enhance the realism and generalization of the $G_{\theta}$'s outputs. Upon reviewing VSD \cref{eq:vsd}, $\boldsymbol{\epsilon}_{\phi}(\boldsymbol{\hat z_{t}};t,c_{y})$ represents the current estimated gradient direction for $G_{\theta}$'s noisy outputs $\boldsymbol{\hat z_{t}}$, whereas $\boldsymbol{\epsilon}_{\psi}(\boldsymbol{\hat z_{t}};t,c_{y})$ corresponds to the ideal gradient direction guiding towards more realistic outputs. 
The overarching goal of model optimization is to align the suboptimal gradient direction with the superior direction based on pre-trained priors, thus facilitating the optimization of the Student distribution toward that of the Teacher. However, this strategy encounters hurdles, especially in the early training phase: the quality of synthetic latent $\boldsymbol{\hat z_{t}}$ is not high enough for the Teacher Model to provide a precise prediction. 
As illustrated in \cref{fig:gradient-figure}, the Teacher Model struggles to accurately predict the optimization direction for low-quality synthetic latent $\boldsymbol{\hat z_{t}}$ in the early stage, as indicated by a cosine similarity of only $0.2$ to the ideal direction, compared to $0.88$ for high-quality latent $\boldsymbol{z_{t}}$. This problem can lead to severe visual artifacts, as is evident in \cref{fig:fig_visual}(a).

% \noindent
% {\bf Approach.}
A straightforward remedial measure is to employ a mean squared error (MSE) loss to align the synthetic latent with the ideal inputs of the Teacher Model, which are derived from the HQ latent. However, as shown in \cref{fig:fig_visual}(b), this approach has been observed to lead to over-smoothed results \cite{he2022revisiting}. Our strategy, instead, is to align the predictions made by the Teacher Model on both synthetic and HQ latent, thereby encouraging greater consistency between them. The core idea is that for samples drawn from the same distribution, the real scores predicted by the Teacher Model should be close to each other. We refer to this approach as Target Score Matching (TSM):
\begin{equation}
\begin{split}
& \nabla_\theta\mathcal{L}_{\mathrm{TSM}}(\boldsymbol{\hat z },\boldsymbol{z},c_y) \\ & =
  \mathbb{E}_{t,\epsilon}\left[w(t)({\epsilon}_\psi(\boldsymbol{\hat z}_t;t,c_y)-{\epsilon}_\psi(\boldsymbol{z}_t;t,c_y))\frac{\partial\boldsymbol{\hat z}}{\partial\theta}\right]
\label{eq:tsm}
\end{split}
\end{equation}
where the expectation of the gradient is computed across all diffusion timesteps $t \in \{1,\cdots,T\}$ and $\epsilon \sim \mathcal{N}(0,I)$.
\Cref{eq:tsm} encapsulates the optimization loss for our Target Score Matching. 
Upon examining it in conjunction with \cref{eq:vsd}, we notice that VSD utilizes the prediction residual between the Teacher and the LoRA Model to drive gradient backpropagation. Similarly, our TSM employs the synthetic and the HQ data to produce the gradients. By blending these two strategies with hyperparameter weights $\lambda$ and $1 - \lambda$, we construct a combined optimization loss that effectively unifies the strengths of both approaches, as formulated in \cref{eq:tsd}, to guide the training process.
\begin{equation}
\begin{split}
\nabla_\theta\mathcal{L}_{\mathrm{TSD}}(\boldsymbol{\hat z},\boldsymbol{z},c_y) = \mathbb{E}_{t,\epsilon}\Big{[} w(t)[{\epsilon}_\psi(\boldsymbol{\hat z}_t;t,c_y)-  \\  {\epsilon}_\psi(\boldsymbol{z}_t;t,c_y) + \lambda({\epsilon}_\psi(\boldsymbol{z}_t;t,c_y) - {\epsilon}_\phi(\boldsymbol{\hat z}_t;t,c_y))]\frac{\partial\boldsymbol{\hat z}}{\partial\theta} \Big{]}
\label{eq:tsd}
\end{split}
\end{equation}
where $w(t)$ is a time-aware weighting function tailored for Real-ISR. 
% Other symbols are in accordance with those previously mentioned. By introducing the prediction of the pre-trained diffusion model on HQ latent, we have circumvented the issue of the model falling into the visual artifacts or over-smoothed problem, as illustrated in \cref{fig:fig_visual}(c).
Other symbols are consistent with those previously defined. By introducing the prediction of the pre-trained diffusion model on HQ latent, we have circumvented the issue of the model falling into visual artifacts or producing over-smoothed results, as illustrated in \cref{fig:fig_visual}(c).

%-------------------------------------------------------------------------
% Distribution-Aware Sampling Module
\subsection{Distribution-Aware Sampling Module} 
\label{subsec:DASM}
\begin{figure}[!t]
  \centering
    \includegraphics[width=\linewidth]{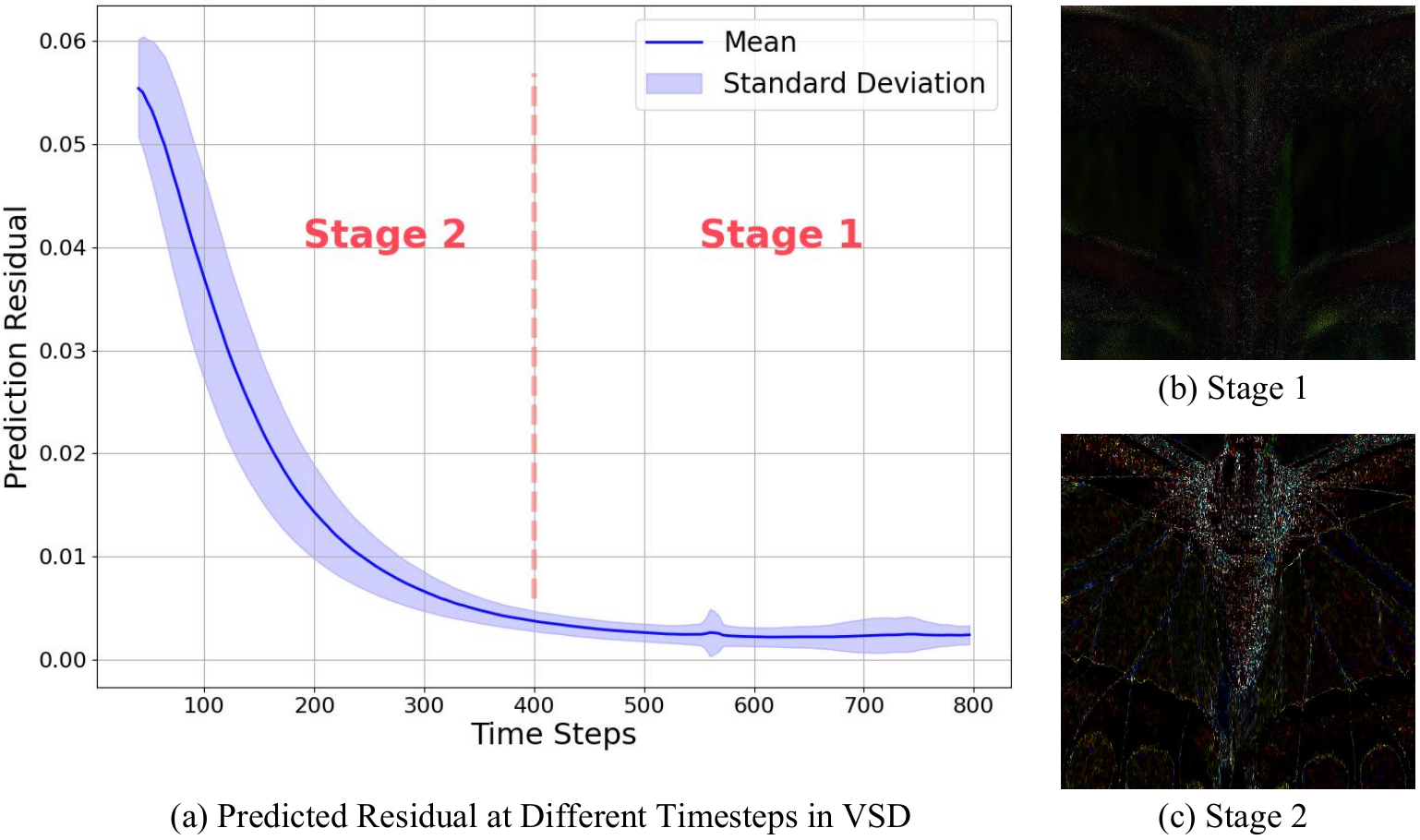}
   \caption{(a) The prediction errors of the VSD loss at different timesteps. The error divergence is more pronounced in early timesteps than later.  This phenomenon is observed throughout the optimization process. (b) The visualization of Stage 1 prediction error. (c) The visualization of Stage 2 prediction error.}
   \label{fig:show}
   \vspace{0cm}
\end{figure}

In the VSD-based framework, it is necessary to match the score functions predicted by the Teacher Model and the LoRA Model across timesteps $t \in {0, 1, \dots, T}$. However, for the Real-ISR problem, this matching performance is inconsistent across timesteps, as illustrated in \cref{fig:show}(a). This phenomenon may be attributed to the reliance on low-frequency (LF) priors in the LQ data, while lacking guidance from high-frequency (HF) details.
The output sample $\boldsymbol{\hat z_0}$, derived from LQ data, contains low-frequency (LF) priors that are easily captured by the LoRA Model, resulting in similar predictions during LF restoration (Stage 1), as shown in \cref{fig:show}(b). However, in Stage 2, due to the absence of high-frequency (HF) details in $\boldsymbol{\hat z_0}$, the LoRA Model struggles to reconstruct fine-grained features, leading to divergent predictions, as illustrated in \cref{fig:show}(c). To address this issue, we aim to reduce such divergence.

\begin{figure}[!t]
  \centering
  \setlength{\abovecaptionskip}{0.1cm}
    \includegraphics[width=\linewidth]{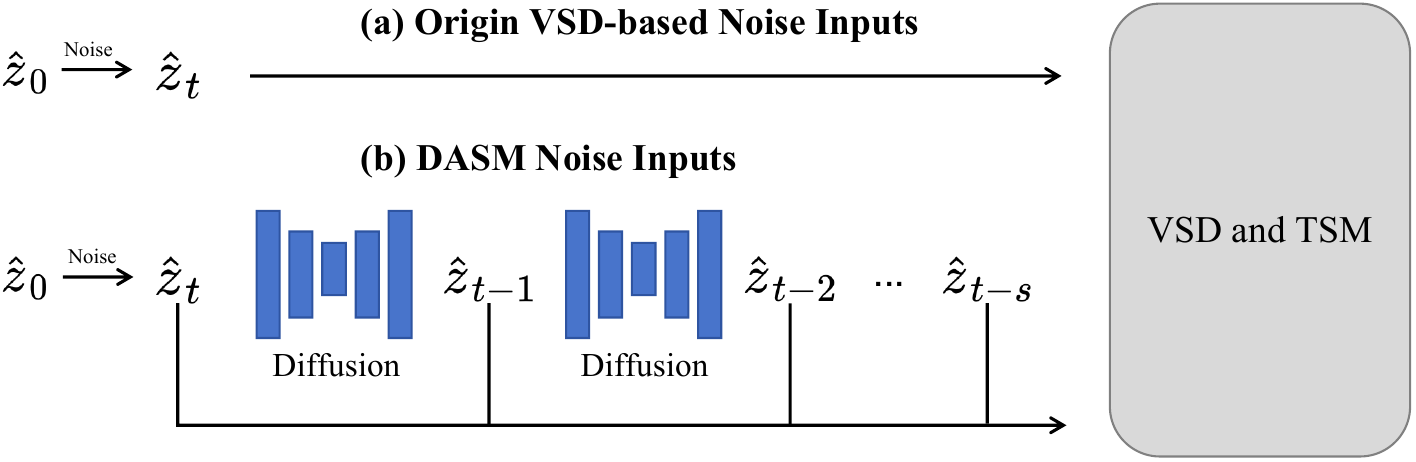}
   \caption{
    Illustration of DASM. Top: The naive approach that adds noise directly to the samples. 
    Bottom: The proposed DASM leverages diffusion model priors to generate noisy latent that better align with the true sampling trajectory. These noisy samples can all serve as inputs to the downstream network, enabling effective gradient backpropagation.
   }
   \label{fig:DAMS}
   \vspace{-0.2cm}
\end{figure}

Existing methods match the score function at each iteration using a single latent sample $\boldsymbol{\hat z_t}$, with the timestep $t$ drawn from a uniform distribution. 
This leads to slow convergence and even training difficulty during Stage 2, as gradients from important timesteps are diluted by uniform averaging.
To this end, we propose our Distribution-Aware Sampling Module (DASM). 
This module accumulates optimization gradients for earlier timestep samples in a single iteration, enabling the backpropagation of more gradients focused on detail optimization. 
As shown in \cref{fig:DAMS}, we first obtain the noisy synthetic latent representation as
 $\boldsymbol{\hat  z_t} = (1 - \sigma_t) \boldsymbol{\hat z_0 }+ \sigma_t \boldsymbol{\epsilon} $, where $\sigma_t$ is a weighting factor and $\epsilon$ denotes Gaussian noise. Subsequently, we employ a LoRA Model to perform denoising, yielding noisy samples at the previous timestep as described in \cref{eq:sample}:
\begin{equation}
\boldsymbol{\hat z_{t-1}} = \boldsymbol{\hat z_t} + ( \sigma_{t-1} - \sigma_t )\cdot \boldsymbol{\epsilon}_\phi(\boldsymbol{\hat z_t};t,c_y),
\label{eq:sample}
\end{equation}
\noindent
% where $\sigma_{t-1}$ and  $\sigma_{t}$ are from the flow matching scheduler, and here LoRA Model has fit $\boldsymbol{\hat  z_0}$'s distribution. Similarly, $\boldsymbol{  z_{t-1}}$ can be obtained by denoising on the Teacher Model.
% Ultimately, in a single iteration, the gradients from noise samples along its trajectory can be accumulated to update the Student Model. Since the obtained samples follow the diffusion sampling trajectory and are directed towards early timestep, we effectively optimize the divergence in Stage 2.
The parameters $\sigma_{t-1}$ and $\sigma_t$ are obtained from the flow matching scheduler. Here, the LoRA Model has learned the distribution of $\boldsymbol{\hat z_0}$. Similarly, $\boldsymbol{z_{t-1}}$ can be obtained by denoising using the Teacher Model.
In a single iteration, gradients from noisy samples along the sampling trajectory can be accumulated to update the Student Model. Since these samples follow the diffusion sampling trajectory and are concentrated at early timesteps, this approach effectively reduces the divergence observed in Stage 2.

%-------------------------------------------------------------------------
% Train Loss
\subsection{Training Objective} 
\label{subsec:overall}
We summarize all the losses that we used in our framework.

\noindent
{\bf Student Model $G_\theta$.}
We train our Student Model with the reconstruction loss $\mathcal{L}_{Rec}$ and the regularization loss $\mathcal{L}_{Reg}$.

\noindent
For the reconstruction loss, we use the $LPIPS$ loss in the pixel space and the $MSE$ loss in the latent space: 
\begin{equation}
\begin{split}
& \mathcal{L}_{Rec}\left(G_\theta(\boldsymbol{x}_L),\boldsymbol{x}_H\right) \\
& = \gamma_1 \mathcal{L}_{LPIPS}\left(G_\theta(\boldsymbol{x}_L),\boldsymbol{x}_H\right) + \mathcal{L}_{MSE}\left(\boldsymbol{z}_t,\boldsymbol{ \hat z}_t\right). 
\end{split}
\label{eq:rec}
\end{equation}
\noindent
For the regularization loss, we use our TSD loss, \cref{eq:tsd}.
% \begin{equation}
% \mathcal{L}_{reg}(q_{\theta}(\hat{x}_H),p(x_H))= \mathcal{L}_{\mathrm{TSD}}(\mathbf{\hat z },\mathbf{z},c_y).
% \label{eq:reg}
% \end{equation}
\noindent
Therefore, the overall training objective for the Student Model $G_\theta$ is:
\begin{equation}
\mathcal{L}_{Stu} = \mathcal{L}_{Rec} + \gamma_2 \mathcal{L}_{Reg},
\label{eq:total}
\end{equation}
where $\gamma_1$ and $\gamma_2$ are weighting factors. We initialize both $\gamma_1$ and $\gamma_2$ to 1 at the beginning of training. As optimization progresses, we ramp up $\gamma_1$  from 1 to 2 while maintaining $\gamma_2$ at its initial value.

\noindent
{\bf LoRA Model $\boldsymbol{\epsilon}_\phi$.}
As stipulated by VSD, the replica $\epsilon_\phi$ must be trainable, with its training objective being:
\begin{equation}
\mathcal{L}_{Diff}(\boldsymbol{{\hat z}},c_y)=\mathbb{E}_{t,\boldsymbol{\epsilon}}[\left\| \boldsymbol{\epsilon}_\phi(\boldsymbol{\hat  z_t};t,c_y)-\boldsymbol{\epsilon}'\right\|^2],
\label{eq:LoRA}
\end{equation}

\noindent
 where $\boldsymbol{\epsilon}'$ serves as the training target for the denoising network, representing Gaussian noise in the context of DDPM, and a gradient towards HQ data for flow matching.
\section{Experiments}
\label{sec:experiment}

\subsection{Experimental Settings}

\noindent
{\bf Training Datasets.}
For training, we utilize DIV2K \cite{agustsson2017ntire}, Flickr2K \cite{timofte2017ntire}, LSDIR \cite{li2023lsdir}, and the first 10K face images from FFHQ \cite{karras2019style}. To synthesize LR-HR pairs, we adopt the same degradation pipeline as in Real-ESRGAN \cite{wang2021real}.

\noindent
{\bf Test Datasets.}
We evaluate our model on the synthetic DIV2K-Val \cite{agustsson2017ntire} dataset, as well as two real-world datasets: RealSR \cite{cai2019toward} and DRealSR \cite{wei2020component}. The real-world datasets consist of 128×128 low-quality (LQ) and 512×512 high-quality (HQ) image pairs. For the synthetic set, 3,000 pairs were generated by cropping 512×512 patches from DIV2K-Val and applying the Real-ESRGAN \cite{wang2021real} degradation pipeline to downsample them to 128×128.
      
\noindent
{\bf Evaluation Metrics.}
To evaluate our method, we employ both full-reference and no-reference metrics. The full-reference metrics include PSNR and SSIM \cite{wang2004image} (computed on the Y channel of the YCbCr color space) for fidelity; LPIPS \cite{zhang2018unreasonable} and DISTS \cite{ding2020image} for perceptual quality; and FID \cite{heusel2017gans} for measuring distribution similarity. The no-reference metrics include NIQE \cite{zhang2015feature}, MANIQA \cite{yang2022maniqa}, MUSIQ \cite{ke2021musiq}, and CLIPIQA \cite{wang2023exploring}.

\noindent
{\bf Compared Methods.}
We categorize the test models into two groups: single-step and multi-step inference. The single-step inference diffusion models include SinSR \cite{wang2024sinsr}, AddSR \cite{xie2024addsr}, and OSEDiff \cite{wu2024one}. The multi-step inference diffusion models comprise StableSR \cite{wang2024exploiting}, ResShift \cite{yue2024resshift}, PASD \cite{yang2023pixel}, DiffBIR \cite{lin2023diffbir}, SeeSR \cite{wu2024seesr}, SUPIR \cite{yu2024scaling}, and AddSR \cite{xie2024addsr}. Specifically, for AddSR, we have conducted comparisons between its single-step and four-step models. GAN-based Real-ISR methods \cite{zhang2021designing,chen2022real,liang2022details,wang2021real} are detailed in the supplementary material.

\noindent
{\bf Implementation Details.}
% All models mentioned are initialized from the Teacher Model (SD3 \cite{esser2024scaling} in our work). Similar to OSEDiff \cite{wu2024one}, we only train the vae encoder and denoising network in the Student Model, frozen the vae decoder in order to preserve vae's prior \cite{kingma2013auto}. We utilize the default prompt for the Student Model, and prompts are extracted from HQ for the Teacher and LoRA models when training. We use AdamW optimizer \cite{loshchilov2017decoupled} with a learning rate of 5e-6 for the Student Model and 1e-6 for the LoRA Model, and set the rank of LoRA to 64 for both two models. During the initial training phase, we incorporate MSE loss in the latent space and exclude DASM to enhance model stability and reduce training time. In later stages, we remove this MSE loss as it leads to over-smoothed results, and we employ DASM to improve image restoration.  The training process took roughly 96h, utilizing 8 NVIDIA V100 GPUs with a batch size of 16. 
All models are initialized from the Teacher Model (SD3 \cite{esser2024scaling} in our work). Similar to OSEDiff \cite{wu2024one}, we only train the VAE encoder and the denoising network in the Student Model, freezing the VAE decoder to preserve its prior \cite{kingma2013auto}.
We utilize the default prompt for the Student Model, while prompts are extracted from HQ images for the Teacher and LoRA models during training. We adopt the AdamW optimizer \cite{loshchilov2017decoupled} with a learning rate of 5e-6 for the Student Model and 1e-6 for the LoRA Model, setting the LoRA rank to 64 for both models.
During the initial training phase, we incorporate MSE loss in the latent space and exclude DASM to stabilize training and reduce time cost. In later stages, we remove the MSE loss to avoid over-smoothed results and introduce DASM to enhance restoration quality.
The training process took approximately 96 hours, utilizing 8 NVIDIA V100 GPUs with a batch size of 16.

%Quantitative -------------------------------------------------------------------------

\begin{table*}[!t]
\setlength{\abovecaptionskip}{0.1cm}
\caption{
Quantitative comparison with the state-of-the-art one-step methods across both synthetic and real-world benchmarks. The number of diffusion inference steps is indicated by ‘s’. The best  results of each metric are highlighted in \textcolor{red}{red}.}
\label{tab:one-step}
\resizebox{\textwidth}{!}{%
\begin{tabular}{@{}c|c|ccccccccc@{}}
\toprule
\textbf{Datasets} & \textbf{Method} & \textbf{PSNR $\uparrow$} & \textbf{SSIM $\uparrow$} & \textbf{LPIPS $\downarrow$ } & \textbf{DISTS $\downarrow$} & \textbf{FID $\downarrow$} & \textbf{NIQE $\downarrow$} & \textbf{MUSIQ $\uparrow$} & \textbf{MANIQA $\uparrow$} & \textbf{CLIPIQA $\uparrow$} \\ \midrule
 & OSEDiff-1s & 27.92 &  {\color[HTML]{D83931} \textbf{0.7836}} & 0.2968 & 0.2162 & 135.51 & 6.4471 & 64.69 & {\color[HTML]{D83931} \textbf{0.5898}} & 0.6958 \\
 & AddSR-1s & 27.77 & 0.7722 & 0.3196 & 0.2242 & 150.18 & 6.9321 & 60.85 & 0.5490 & 0.6188 \\
 & SinSR-1s & {\color[HTML]{D83931} \textbf{28.38}} & 0.7497 & 0.3669 & 0.2484 & 172.72 & 6.9606 & 55.03 & 0.4904 & 0.6412 \\
\multirow{-4}{*}{\textbf{DRealSR}} & Ours-1s & 27.77 & 0.7559 & {\color[HTML]{D83931} \textbf{0.2967}} & {\color[HTML]{D83931} \textbf{0.2136}} & {\color[HTML]{D83931} \textbf{134.98}} & {\color[HTML]{D83931} \textbf{5.9131}} & {\color[HTML]{D83931} \textbf{66.62}} & 0.5874 & {\color[HTML]{D83931} \textbf{0.7344}} \\ \midrule
 & OSEDiff-1s & 25.15 & 0.7341 & 0.2920 & 0.2128 & 123.48 & 5.6471 & 69.10 & 0.6326 & 0.6687 \\
 & AddSR-1s & 24.79 & 0.7077 & 0.3091 & 0.2191 & 132.05 & 5.5440 & 66.18 & 0.6098 & 0.5722 \\
 & SinSR-1s & {\color[HTML]{D83931} \textbf{26.27}} & {\color[HTML]{D83931} \textbf{0.7351}} & 0.3217 & 0.2341 & 137.59 & 6.2964 & 60.76 & 0.5418 & 0.6163 \\
\multirow{-4}{*}{\textbf{RealSR}} & Ours-1s & 24.81 & 0.7172 & {\color[HTML]{D83931} \textbf{0.2743}} & {\color[HTML]{D83931} \textbf{0.2104}} & {\color[HTML]{D83931} \textbf{114.45}} & {\color[HTML]{D83931} \textbf{5.1298}} & {\color[HTML]{D83931} \textbf{71.19}} & {\color[HTML]{D83931} \textbf{0.6347}} & {\color[HTML]{D83931} \textbf{0.7160}} \\ \midrule
 & OSEDiff-1s & 23.72 & {\color[HTML]{D83931} \textbf{0.6109}} & 0.2942 & 0.1975 & {\color[HTML]{D83931} \textbf{26.34}} & 4.7089 & 67.96 & 0.6131 & 0.6681 \\
 & AddSR-1s & 23.26 & 0.5902 & 0.3623 & 0.2123 & 29.68 & 4.7610 & 63.39 & 0.5657 & 0.5734 \\
 & SinSR-1s & {\color[HTML]{D83931} \textbf{24.41}} & 0.6018 & 0.3262 & 0.2068 & 35.55 & 5.9981 & 62.95 & 0.5430 & 0.6501 \\
\multirow{-4}{*}{\textbf{DIV2K-Val}} & Ours-1s & 23.02 & 0.5808 & {\color[HTML]{D83931} \textbf{0.2673}} & {\color[HTML]{D83931} \textbf{0.1821}} & 29.16 & {\color[HTML]{D83931} \textbf{4.3244}} & {\color[HTML]{D83931} \textbf{71.69}} & {\color[HTML]{D83931} \textbf{0.6192}} & {\color[HTML]{D83931} \textbf{0.7416}} \\ \bottomrule
\end{tabular}%
}
\end{table*}

\subsection{Comparison with Existing Methods}
\label{subsec:quantitative}
\noindent
% Quantitative Comparisons.
{\bf Quantitative Comparisons.}
% \cref{tab:one-step} shows the quantitative comparison of our method with single-step  diffusion models on three datasets. Ours achieves the best results on most evaluation metrics. SinSR and AddSR, as distilled versions of previous multi-step super-resolution methods, reduce inference steps but experience a corresponding decrease in performance metrics. OSEDiff introduces the VSD loss from 3D generation tasks into Real-ISR without fully addressing the substantial differences between the two domains. Therefore, the no-reference metrics for image quality are not very satisfied. In contrast, our proposed TSD-SR, tailored for Real-ISR, outperforms other single-step models in terms of the vast majority of key metrics.
\cref{tab:one-step} shows the quantitative comparison of our method with other single-step diffusion models on three datasets. Our method achieves the best results across most evaluation metrics.
SinSR and AddSR, as distilled versions of previous multi-step super-resolution methods, significantly reduce inference steps while suffering a corresponding drop in performance. 
OSEDiff introduces the VSD loss from 3D generation tasks into the Real-ISR without fully accounting for the substantial differences between the two domains. As a result, its no-reference image quality metrics are not satisfactory.
In contrast, our proposed TSD-SR, specifically designed for Real-ISR, outperforms all other single-step models in the vast majority of key metrics.

% \cref{tab:multi-step} show the quantitative comparison with multi-step models. We can draw the following conclusions:
% (1) TSD-SR demonstrates significant advantages over competing methods in terms of LPIPS, DISTS, and NIQE metrics. Additionally, it achieves performance that surpasses most multi-step models on FID, MUSIQ, and CLIPIQA. 
% (2) DiffBIR, SeeSR, PASD, and AddSR exhibit better performance on the MANIQA metric, which may be attributed to the fact that multi-step models have more denoising iterations to produce rich details.
% (3) ResShift stands out with the highest PSNR and SSIM scores, while StableSR also shows notable performance regarding DISTS and FID metrics. However, both two models underperform in terms of no-reference metrics.
\cref{tab:multi-step} shows the quantitative comparison with multi-step models. We can draw the following conclusions:
(1) TSD-SR demonstrates significant advantages over competing methods in terms of LPIPS, DISTS, and NIQE. Additionally, it outperforms most multi-step models in FID, MUSIQ, and CLIPIQA.
(2) DiffBIR, SeeSR, PASD, and AddSR achieve better results in terms of MANIQA, which may be attributed to the fact that multi-step models benefit from more denoising iterations to generate richer details.
(3) ResShift stands out with the highest PSNR and SSIM scores, while StableSR also performs well in terms of DISTS and FID. However, both models underperform on the no-reference metrics.

% Finally, we explain the lower PSNR and SSIM metrics in our experiments. Several works \cite{xie2024addsr, yu2024scaling} have found that these reconstruction metrics  are not well-suited for evaluating Real-ISR tasks. When a model recovers better details, it leads to lower reconstruction metrics, indicating a trade-off \cite{blau2018perception,zhang2022perception,luo2024skipdiff}. This phenomenon has also been widely discussed in other related research work \cite{blau2018perception,zhang2022perception,luo2024skipdiff,yu2024scaling, zhang2018unreasonable,wang2024exploiting, xie2024addsr}.
Finally, we explain the relatively lower PSNR and SSIM scores observed in our experiments. Several studies \cite{xie2024addsr, yu2024scaling} have shown that these reconstruction metrics are not well-suited for the evaluation of Real-ISR tasks. Models that recover more realistic or detailed textures often yield lower PSNR and SSIM scores, reflecting a fundamental trade-off between perceptual quality and pixel-wise fidelity \cite{blau2018perception,zhang2022perception,luo2024skipdiff}. This phenomenon has also been extensively discussed in the other research work \cite{blau2018perception,zhang2022perception,luo2024skipdiff,yu2024scaling,zhang2018unreasonable,wang2024exploiting,xie2024addsr}.
LPIPS \cite{zhang2018unreasonable} is proposed to overcome the limitation that PSNR and SSIM fail to align with human judgments in spatial ambiguities situations.
Other DMs-based SR researchers \cite{yu2024scaling, wang2024exploiting} argue that DMs introduce superior pre-trained priors, enabling the restoration of information that traditional methods (from scratch) cannot achieve. However, such capability often leads to a decline in pixel-level metrics, as they prioritize distribution modeling and sampling from learned distributions over strict pixel fidelity.
We anticipate the development of better full-reference metrics in the future to assess advanced Real-ISR methods. 
Refer to the supplementary material for detailed visual comparisons.

\begin{table*}[!t]
\setlength{\abovecaptionskip}{0.1cm}
\caption{Quantitative comparison with state-of-the-art multi-step methods across both synthetic and real-world benchmarks. The number of diffusion inference steps is indicated by ‘s’. The best and second best results of each metric are highlighted in \textcolor{red}{red} and \textcolor{blue}{blue}, respectively.}
\label{tab:multi-step}
\resizebox{\textwidth}{!}{%
\begin{tabular}{@{}c|c|ccccccccc@{}}
\toprule
\textbf{Datasets} & \textbf{Method} & \textbf{PSNR $\uparrow$} & \textbf{SSIM $\uparrow$} & \textbf{LPIPS $\downarrow$ } & \textbf{DISTS $\downarrow$} & \textbf{FID $\downarrow$} & \textbf{NIQE $\downarrow$} & \textbf{MUSIQ $\uparrow$} & \textbf{MANIQA $\uparrow$} & \textbf{CLIPIQA $\uparrow$} \\ \midrule
  & StableSR-200s & 28.04 & 0.7454 & 0.3279 & {\color[HTML]{4E83FD} \textbf{0.2272}} & {\color[HTML]{4E83FD} \textbf{144.15}} & 6.5999 & 58.53 & 0.5603 & 0.6250 \\
 & DiffBIR-50s & 25.93 & 0.6525 & 0.4518 & 0.2761 & 177.04 & {\color[HTML]{4E83FD} \textbf{6.2324}} & 65.66 & {\color[HTML]{D83931} \textbf{0.6296}} & 0.6860 \\
 & SeeSR-50s & {\color[HTML]{4E83FD} \textbf{28.14}} & {\color[HTML]{4E83FD} \textbf{0.7712}} & {\color[HTML]{4E83FD} \textbf{0.3141}} & 0.2297 & 146.95 & 6.4632 & 64.74 & 0.6022 & 0.6893 \\
 & SUPIR-50s & 25.09 & 0.6460 & 0.4243 & 0.2795 & 169.48 & 7.3918 & 58.79 & 0.5471 & 0.6749 \\
 & PASD-20s & 27.79 & 0.7495 & 0.3579 & 0.2524 & 171.03 & 6.7661 & 63.23 & 0.5919 & 0.6242 \\
 & ResShift-15s & {\color[HTML]{D83931} \textbf{28.69}} & {\color[HTML]{D83931} \textbf{0.7874}} & 0.3525 & 0.2541 & 176.77 & 7.8762 & 52.40 & 0.4756 & 0.5413 \\
 & AddSR-4s & 26.72 & 0.7124 & 0.3982 & 0.2711 & 164.12 & 7.6689 & {\color[HTML]{4E83FD} \textbf{66.33}} & {\color[HTML]{4E83FD} \textbf{0.6257}} & {\color[HTML]{4E83FD} \textbf{0.7226}} \\
\multirow{-8}{*}{\textbf{DRealSR}} & Ours-1s & 27.77 & 0.7559 & {\color[HTML]{D83931} \textbf{0.2967}} & {\color[HTML]{D83931} \textbf{0.2136}} & {\color[HTML]{D83931} \textbf{134.98}} & {\color[HTML]{D83931} \textbf{5.9131}} & {\color[HTML]{D83931} \textbf{66.62}} & 0.5874 & {\color[HTML]{D83931} \textbf{0.7344}} \\ \midrule
 & StableSR-200s & 24.62 & 0.7041 & 0.3070 & {\color[HTML]{4E83FD} \textbf{0.2156}} & 128.54 & 5.7817 & 65.48 & 0.6223 & 0.6198 \\
 & DiffBIR-50s & 24.24 & 0.6650 & 0.3469 & 0.2300 & 134.56 & 5.4932 & 68.35 & {\color[HTML]{4E83FD} \textbf{0.6544}} & 0.6961 \\
 & SeeSR-50s & 25.21 & 0.7216 & {\color[HTML]{4E83FD} \textbf{0.3003}} & 0.2218 & {\color[HTML]{4E83FD} \textbf{125.10}} & {\color[HTML]{4E83FD} \textbf{5.3978}} & 69.69 & 0.6443 & 0.6671 \\
 & SUPIR-50s & 23.65 & 0.6620 & 0.3541 & 0.2488 & 130.38 & 6.1099 & 62.09 & 0.5780 & 0.6707 \\
 & PASD-20s & {\color[HTML]{4E83FD} \textbf{25.68}} & {\color[HTML]{4E83FD} \textbf{0.7273}} & 0.3144 & 0.2304 & 134.18 & 5.7616 & 68.33 & 0.6323 & 0.5783 \\
 & ResShift-15s & {\color[HTML]{D83931} \textbf{26.39}} & {\color[HTML]{D83931} \textbf{0.7567}} & 0.3158 & 0.2432 & 149.59 & 6.8746 & 60.22 & 0.5419 & 0.5496 \\
 & AddSR-4s & 23.33 & 0.6400 & 0.3925 & 0.2626 & 154.22 & 5.8959 & {\color[HTML]{D83931} \textbf{71.49}} & {\color[HTML]{D83931} \textbf{0.6826}} & {\color[HTML]{D83931} \textbf{0.7225}} \\
\multirow{-8}{*}{\textbf{RealSR}} & Ours-1s & 24.81 & 0.7172 & {\color[HTML]{D83931} \textbf{0.2743}} & {\color[HTML]{D83931} \textbf{0.2104}} & {\color[HTML]{D83931} \textbf{114.45}} & {\color[HTML]{D83931} \textbf{5.1298}} & {\color[HTML]{4E83FD} \textbf{71.19}} & 0.6347 & {\color[HTML]{4E83FD} \textbf{0.7160}} \\ \midrule
 & StableSR-200s & 23.27 & 0.5722 & {\color[HTML]{4E83FD} \textbf{0.3111}} & 0.2046 & {\color[HTML]{D83931} \textbf{24.95}} & 4.7737 & 65.78 & 0.6164 & 0.6753 \\
 & DiffBIR-50s & 23.13 & 0.5717 & 0.3469 & 0.2108 & 33.93 & {\color[HTML]{4E83FD} \textbf{4.6056}} & 68.54 & {\color[HTML]{4E83FD} \textbf{0.6360}} & 0.7125 \\
 & SeeSR-50s & 23.73 & {\color[HTML]{4E83FD} \textbf{0.6057}} & 0.3198 & {\color[HTML]{4E83FD} \textbf{0.1953}} & {\color[HTML]{4E83FD} \textbf{25.81}} & 4.8322 & 68.49 & 0.6198 & 0.6899 \\
 & SUPIR-50s & 22.13 & 0.5279 & 0.3919 & 0.2312 & 31.40 & 5.6767 & 63.86 & 0.5903 & 0.7146 \\
 & PASD-20s & {\color[HTML]{4E83FD} \textbf{24.00}} & 0.6041 & 0.3779 & 0.2305 & 39.12 & 4.8587 & 67.36 & 0.6121 & 0.6327 \\
 & ResShift-15s & {\color[HTML]{D83931} \textbf{24.71}} & {\color[HTML]{D83931} \textbf{0.6234}} & 0.3473 & 0.2253 & 42.01 & 6.3615 & 60.63 & 0.5283 & 0.5962 \\
 & AddSR-4s & 22.16 & 0.5280 & 0.4053 & 0.2360 & 35.41 & 5.2584 & {\color[HTML]{4E83FD} \textbf{70.99}} & {\color[HTML]{D83931} \textbf{0.6596}} & {\color[HTML]{D83931} \textbf{0.7593}} \\
\multirow{-8}{*}{\textbf{DIV2K-Val}} & Ours-1s & 23.02 & 0.5808 & {\color[HTML]{D83931} \textbf{0.2673}} & {\color[HTML]{D83931} \textbf{0.1821}} & 29.16 & {\color[HTML]{D83931} \textbf{4.3244}} & {\color[HTML]{D83931} \textbf{71.69}} & 0.6192 & {\color[HTML]{4E83FD} \textbf{0.7416}} \\ \bottomrule
\end{tabular}%
}
\end{table*}

\begin{figure*}[!htbp]
  \centering
  \setlength{\abovecaptionskip}{0.1cm}
    \begin{subfigure}{\linewidth}
    \includegraphics[width=\linewidth]{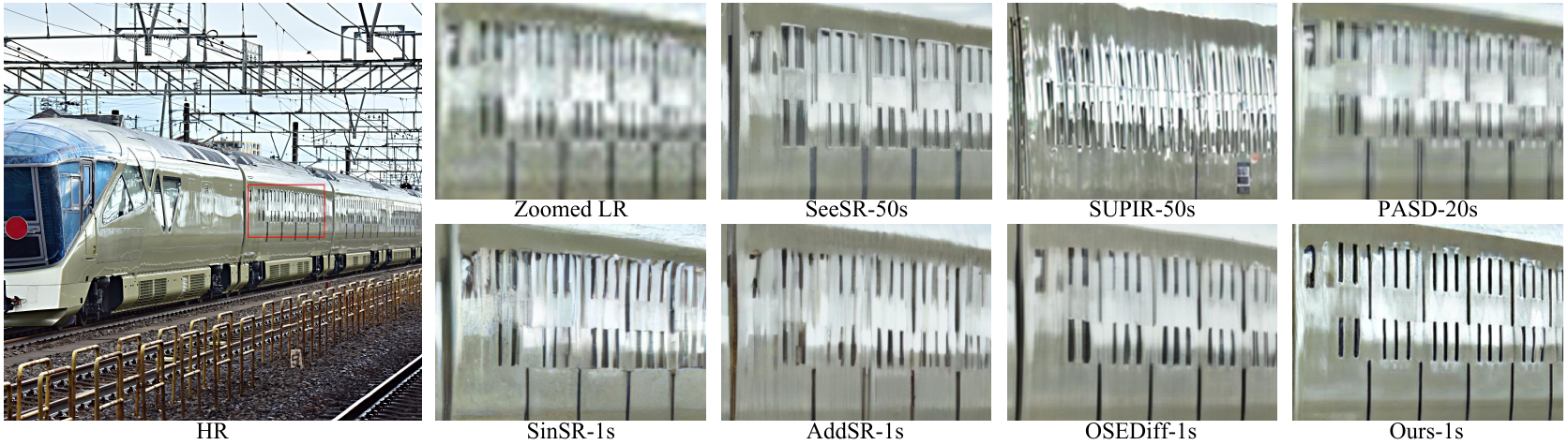}
  \end{subfigure}
    \begin{subfigure}{\linewidth}
    \includegraphics[width=\linewidth]{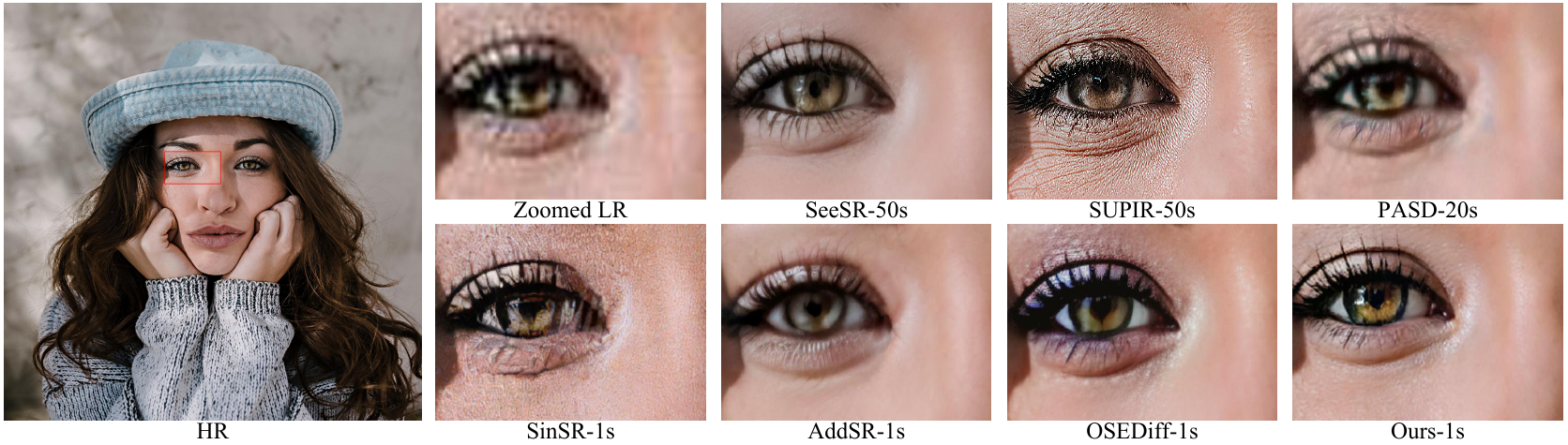}
  \end{subfigure}
   \caption{Visual comparisons of different Real-ISR methods. Please zoom in for a better view.}
   \label{fig:visualization1}
   % \vspace{-0.5cm}
\end{figure*}

% Qualitative -------------------------------------------------------------------------
\noindent
{\bf Qualitative Comparisons.}
\label{subsec:qualitative}
% Qualitative Comparisons.
% \cref{fig:visualization1} presents visual comparisons of different Real-ISR methods. As shown in the multi-step methods' results, SeeSR uses degradation-aware semantic cues to leverage image generation priors, but it sometimes produces over-smoothed textures. 
\cref{fig:visualization1} presents visual comparisons of different Real-ISR methods. As shown in the results of multi-step methods, SeeSR leverages degradation-aware semantic cues to incorporate image generation priors, but it tends to produce over-smoothed textures in some cases.
% SUPIR demonstrates a notably robust generative capacity, however, the overproduction of extraneous details will lead to results that lack naturalism in image restoration (\eg Adding excessive wrinkles at the corners of a young girl's eyes). 
SUPIR demonstrates notably robust generative capabilities. However, the excessive generation of fine details can result in outputs that appear less natural (e.g., adding unnecessary wrinkles around the eyes of a young girl).
% Under more realistic degradation conditions, PASD finds it challenging to restore the corresponding content, revealing a deficiency in the model's robustness capabilities.
% Among single-step methods, SinSR introduces artifacts, possibly because its distilled pre-trained model is trained from scratch, lacking sufficient real-world priors, which results in unsatisfactory image restoration capabilities.
Under more realistic degradation conditions, PASD finds it difficult to recover the appropriate content, indicating limited robustness.
% Among single-step methods, SinSR introduces artifacts, possibly because its base model ResShift is trained from scratch, lacking sufficient real-world priors, which results in unsatisfactory image restoration capabilities.
Among single-step methods, SinSR tends to produce artifacts, likely due to its base model, ResShift, being trained from scratch without adequate exposure to real-world priors, which leads to inferior image restoration quality.
AddSR produces over-smoothed results when using its 1-step model.
% OSEDiff offers improved restoration effects compared to SinSR and AddSR, yet it may fall short in terms of authenticity and naturalness when it comes to detail recovery.
% In contrast, our method effectively generates rich textures and realistic details with enhanced sharpness and contrast. Additional visual comparisons and results are provided in the {\bf supplementary material}.
OSEDiff demonstrates better restoration performance than SinSR and AddSR; however, it may fall short in terms of authenticity and naturalness, particularly in recovering fine details.
In contrast, our method effectively generates rich textures and realistic details with enhanced sharpness and contrast. Additional visual comparisons and results are provided in the supplementary material.

% \begin{figure*}[!htbp]
%   \centering
%   \setlength{\abovecaptionskip}{0.1cm}
%   \begin{subfigure}{\linewidth}
%     \includegraphics[width=\linewidth]{visualization-real_1.pdf}
%   \end{subfigure}
%     \begin{subfigure}{\linewidth}
%     \includegraphics[width=\linewidth]{visualization-real_2.pdf}
%   \end{subfigure}
%    \caption{Visual comparisons on real-world LR images. Please zoom in for a better view.}
%    \label{fig:visualization2}
%    \vspace{-0.5cm}
% \end{figure*}

% Complexity -------------------------------------------------------------------------
\noindent
{\bf Complexity Comparisons}
\label{subsec:complexity}
\begin{table*}[]
\setlength{\abovecaptionskip}{0.1cm}
\caption{Comparison of computational complexity across different diffusion model-based methods. Performance is measured on an A100 GPU using 512×512 input images, excluding model weight and data loading time.}
\label{tab:time}
\resizebox{\textwidth}{!}{%
\begin{tabular}{c|cccccccccccc}
\toprule
\textbf{} & \textbf{StableSR} & \textbf{DiffBIR} & \textbf{SeeSR} & \textbf{SUPIR} & \textbf{PASD} & \textbf{ResShift} & \textbf{AddSR} & \textbf{AddSR} & \textbf{SinSR} & \textbf{OSEDiff} & \textbf{Ours}  \\ \midrule
\textbf{Inference Step} & 200 & 50 & 50 & 50 & 20 & 15 & 4 & 1 & 1 & 1 & 1  \\
\textbf{Inference Time} & 12.4151 & 7.9637 & 5.8167 & 16.8704 & 4.8441 & 0.7546 & 1.0199 & 0.5043 & {\color[HTML]{4E83FD} \textbf{0.1424}} & 0.1650 & {\color[HTML]{D83931} \textbf{0.1362}}  \\ \bottomrule
\end{tabular}%
}
\end{table*}
We assess the computational complexity of the state-of-the-art DM-based Real-ISR methods, as detailed in \cref{tab:time}, with a focus on inference time. Each method is benchmarked on an A100 GPU using input images of size 512×512 pixels.
% We have disregarded the loading time for model weights and data, and the main computation time includes: (1) text extraction time when a text extractor is used; (2) text encoder computation time if needed; (3) VAE encoding and decoding time; (4) denoising network execution time.
% It is evident that TSD-SR has a substantial advantage in inference time compared with multi-step models. Specifically, TSD-SR is more than 120 times faster than SUPIR, 90 times faster than StableSR, about 50 times faster than DiffBIR, more than 40 times faster than SeeSR, more than 35 times faster than PASD, and 4 times faster than ResShift. When compared with existing one-step models, our model boasts the fastest inference times. This is because we directly denoise from LQ data and use a fixed prompt.
We disregarded the loading time for model weights and data. The main computation time consists of: (1) text extraction time (if a text extractor is used); (2) text encoder computation time (if applicable); (3) VAE encoding and decoding time; and (4) denoising network execution time.
It is evident that TSD-SR holds a substantial advantage in inference speed compared to multi-step models. Specifically, TSD-SR is over 120× faster than SUPIR, 90× faster than StableSR, approximately 50× faster than DiffBIR, over 40× faster than SeeSR, more than 35× faster than PASD, and 4× faster than ResShift.
When compared with existing one-step models, our method achieves the fastest inference times. This advantage is attributed to directly denoising from LQ data and employing a fixed prompt.

% User Study ------------------------------------------------
\subsection{User Study}
\label{sec:user-study}
\begin{figure}
  \centering
  \begin{subfigure}{0.48\linewidth}
    \includegraphics[width=\linewidth]{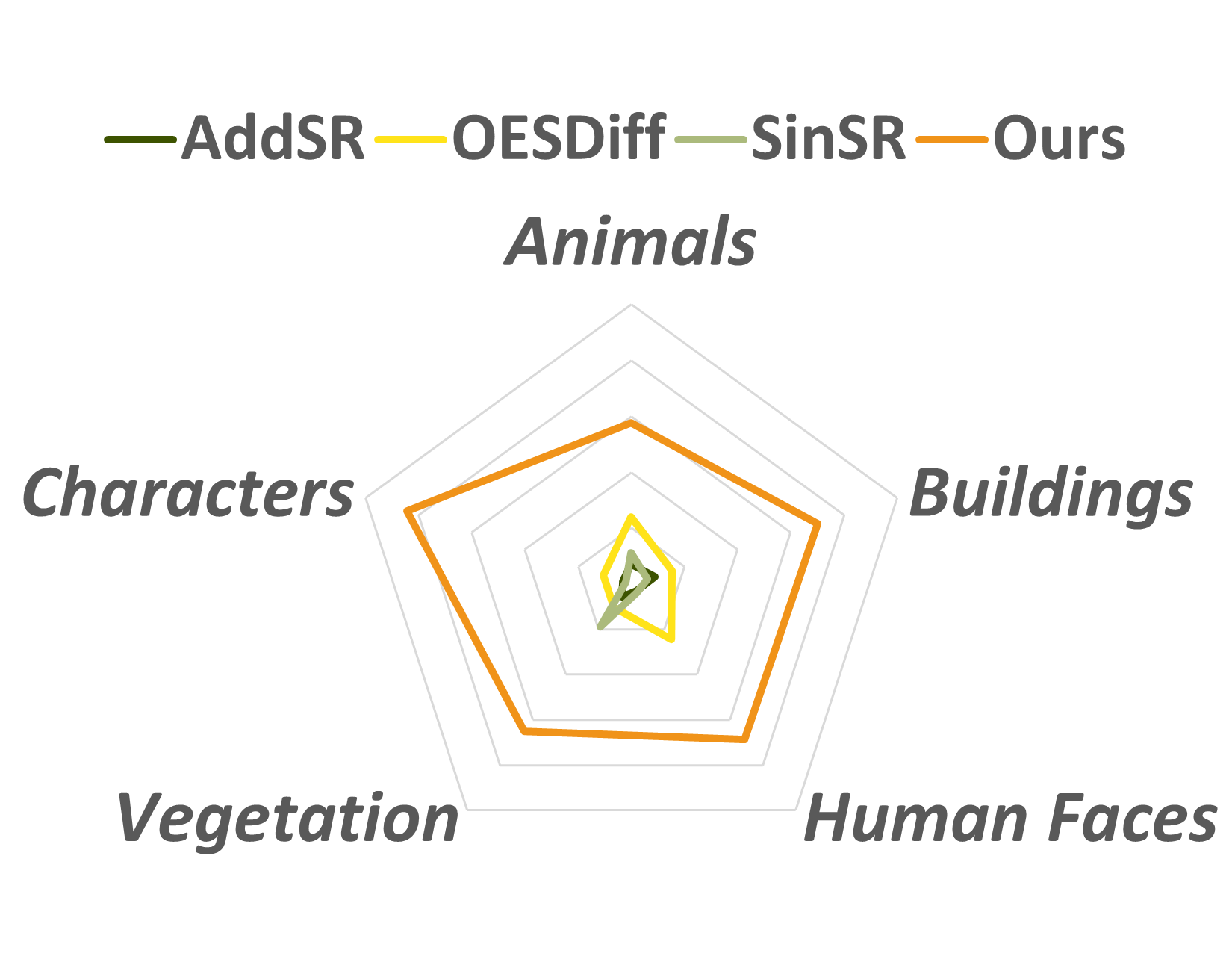}
    \label{fig:latar}
  \end{subfigure}
  \hfill
  \begin{subfigure}{0.48\linewidth }
    \includegraphics[width=\linewidth]{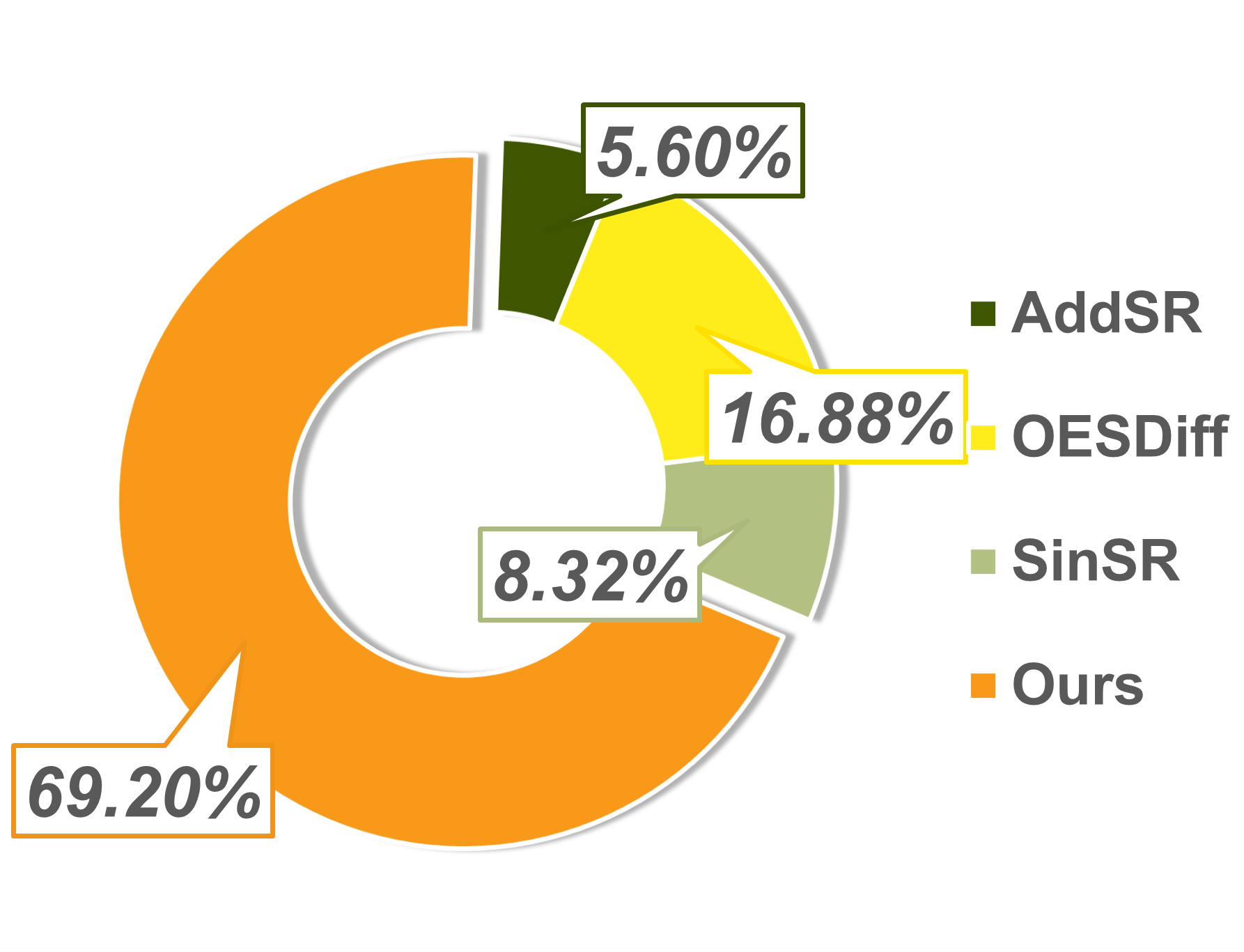}
    \label{fig:pie}
  \end{subfigure}
  \caption{Results of our user study. Left: Category-based user preference radar chart, showing that our model received the highest favor across all categories. Right: User preference pie chart, illustrating that our approach garnered a 69.2\% user satisfaction rating.}
  % \vspace{-0.2cm}
  \label{fig:user-study}
\end{figure}

% We conduct a user study comparing our method with three other diffusion-based one-step super-resolution methods. To make the evaluation more comprehensive, we selected images from five categories—human faces, buildings, animals, vegetation and characters. A total of 50 participants were engaged in the voting process. We guided participants to evaluate the best restoration results based on the similarity to HQ image, structural similarity to the LQ image, and realism of textures and details. The results shown in the \cref{fig:user-study} indicate that our method receives a 69.2\% approval rate from users. We scored 57.60\% in Animals, 70.00\% in Buildings, 68.80\% in Human Faces, 65.20\% in Vegetation, and 84.40\% in Characters, surpassing other models.
We conduct a user study comparing our method with three other diffusion-based one-step super-resolution methods. To ensure a comprehensive evaluation, we selected images from five categories—human faces, buildings, animals, vegetation, and characters. A total of 50 participants took part in the voting process. Participants were instructed to select the best restoration results based on similarity to the HQ image, structural similarity to the LQ image, and realism of textures and details.
The results in \cref{fig:user-study} indicate that our method received a 69.2\% approval rate from users. Specifically, our method achieved 57.6\% in Animals, 70.0\% in Buildings, 68.8\% in Human Faces, 65.2\% in Vegetation, and 84.4\% in Characters, surpassing those of other methods.

% Ablation ------------------------------------------------
\subsection{Ablation Study}
\label{sec:ablation}
{\bf Effectiveness of TSM and DASM.}
% To validate the effectiveness of TSM loss and DASM, we conduct ablation studies by removing them separately in our experiments. We select LPIPS, DISTS, MUSIQ, MANIQA and CLIPIQA for comparison, as these metrics are critical for image quality assessment. Additionally, we select FID as the metric to evaluate distribution similarity. The results are presented in \cref{tab:ablation}. We draw the following conclusions: (1) The absence of TSM loss and DASM negatively impacts performance across both reference-based (LPIPS, DISTS) and non-reference-based metrics (MUSIQ, MANIQA and CLIPIQA). The FID metric is also adversely affected, suggesting a decline in distribution fidelity. (2) The lack of TSM leads to a significant decrease in LPIPS, DISTS, MUSIQ, and CLIPIQA metrics, possibly due to the unreliable directions in VSD causing unrealistic generations. (3) The absence of DASM results in a decline in FID, MUSIQ and CLIPIQA, possibly due to the suboptimal optimization of details.
To validate the effectiveness of the TSM loss and DASM, we conduct ablation studies by removing each component separately. We select LPIPS, DISTS, MUSIQ, MANIQA, and CLIPIQA for comparison, as these metrics are critical for image quality assessment. Additionally, FID is used to evaluate distribution similarity. The results are presented in \cref{tab:ablation}. From the results, we draw the following conclusions:
(1) The absence of TSM loss and DASM negatively impacts performance across both reference-based metrics (LPIPS, DISTS) and no-reference metrics (MUSIQ, MANIQA, and CLIPIQA). The FID metric is also adversely affected, indicating a decline in distribution fidelity.
(2) The lack of TSM leads to a significant decrease in LPIPS, DISTS, MUSIQ, and CLIPIQA, possibly due to unreliable directions in VSD leading to unrealistic generations.
(3) The absence of DASM results in a decline in FID, MUSIQ, and CLIPIQA, possibly due to suboptimal detail optimization.

\begin{table}[!t]
\caption{Ablation study of Target Score Matching loss and Distribution-Aware Sampling Module.}
\label{tab:ablation}
\resizebox{0.47\textwidth}{!}
{%
\begin{tabular}{@{}c|c|cccccc@{}}
\toprule
\textbf{Datasets} & \textbf{Method} & \textbf{LPIPS $\downarrow$} & \textbf{DISTS $\downarrow$} & \textbf{FID $\downarrow$} & \textbf{MUSIQ $\uparrow$} & \textbf{MANIQA $\uparrow$} & \textbf{CLIPIQA $\uparrow$} \\ \midrule
\multirow{3}{*}{\textbf{DRealSR}} & w/o TSM & 0.3176 & 0.2327 & 139.67 & 63.90 & 0.5749 & 0.6958 \\
 & w/o DASM & 0.3097 & 0.2311 & 143.97 & 63.56& 0.5812 & 0.7123 \\
 & Full & {\color[HTML]{D83931} \textbf{0.2967}} & {\color[HTML]{D83931} \textbf{0.2136}} & {\color[HTML]{D83931} \textbf{134.98}} & {\color[HTML]{D83931} \textbf{66.62}} & {\color[HTML]{D83931} \textbf{0.5874}} & {\color[HTML]{D83931} \textbf{0.7344}} \\ 
 \midrule
\multirow{3}{*}{\textbf{RealSR}} & w/o TSM  & 0.2934 & 0.2397 & 117.15 & 68.56 & 0.6338 & 0.6987 \\
 & w/o DASM & 0.2873 & 0.2273 & 123.51 & 69.05 & 0.6273 & 0.7031 \\
 & Full & {\color[HTML]{D83931} \textbf{0.2743}} & {\color[HTML]{D83931} \textbf{0.2104}}  & {\color[HTML]{D83931} \textbf{114.45}}  & {\color[HTML]{D83931} \textbf{71.19}} & {\color[HTML]{D83931} \textbf{0.6347}} & {\color[HTML]{D83931} \textbf{0.7160}} \\ \bottomrule
\end{tabular}%
}
% \vspace{-0.5cm}
\end{table}

\noindent
{\bf Base model for fairer comparison.} 
% To validate the effectiveness of our method across different versions of SD models, we conduct additional experiments, as shown in \cref{tab:re-table2}, including SD2-base and SD2.1-base models. The performance is evaluated on the DRealSR test dataset \cite{wei2020component}. It demonstrates superior performance compared to other one-step SR methods, including OSEDiff \cite{wu2024one} and AddSR \cite{xie2024addsr}. Specifically, our SD2-base version model outperforms the single-step AddSR across all perceptual reference metrics and no-reference metrics, particularly excelling in NIQE\cite{zhang2015feature}, MUSIQ\cite{ke2021musiq}, and CLIPIQA\cite{wang2023exploring}, significantly surpassing the performance of one step AddSR. Our SD2.1-base model demonstrates comparable performance to OSEDiff and surpasses it across various metrics, with notable improvements in NIQE and CLIPIQA.
To validate the effectiveness of our method across different versions of SD models, we conduct additional experiments on SD2-base and SD2.1-base models, as shown in \cref{tab:re-table2}. The performance is evaluated on the DRealSR test dataset \cite{wei2020component}. Our method demonstrates superior performance compared to other one-step SR methods, including OSEDiff \cite{wu2024one} and AddSR \cite{xie2024addsr}. Specifically, our SD2-base model outperforms single-step AddSR across all perceptual reference and no-reference metrics, particularly excelling in NIQE \cite{zhang2015feature}, MUSIQ \cite{ke2021musiq}, and CLIPIQA \cite{wang2023exploring}. Meanwhile, our SD2.1-base model shows comparable or better performance than OSEDiff across various metrics, with notable improvements in NIQE and CLIPIQA.

\newcommand{\redcolor}[1]{{\color[HTML]{D83931} \textbf{#1}}}
\newcommand{\bluecolor}[1]{{\color[HTML]{4E83FD} \textbf{#1}}}
\begin{table}[t]
\centering
\caption{Fair comparison using the same base model to validate TSD-SR}
\label{tab:re-table2}
    \resizebox{0.48\textwidth}{!}{
    \large
    \begin{tabular}{*{10}{c}}
        \toprule
       \textbf{Model} & \textbf{Base Model}  & \textbf{LPIPS$\downarrow$} & \textbf{DISTS$\downarrow$}  & \textbf{NIQE$\downarrow$} & \textbf{MUSIQ$\uparrow$}   &\textbf{CLIPIQA$\uparrow$} \\
        \midrule
        AddSR & SD2-base  & 0.3196 & 0.2242 & 6.9321 & 60.85 & 0.6188 \\
        Ours & SD2-base  & \redcolor{0.3040} & \redcolor{0.2234} & \redcolor{6.2202} & \redcolor{65.14}  & \redcolor{0.6935} \\
        \midrule
        OSEDiff & SD2.1-base  & 0.2968 & 0.2162 & 6.4471 & 64.69  & 0.6958 \\
        Ours & SD2.1-base  & \redcolor{0.2943} & \redcolor{0.2115} & \redcolor{5.7934} & \redcolor{65.41} &\redcolor{0.7109} \\
        % \midrule
        % Ours & SD3 & 26.01 & 0.7174 & 0.3092 & 5.5044 & 66.66 & 0.7345 \\
        \bottomrule
    \end{tabular}
}
    % \vspace{-1.3em}
\end{table}

\noindent
{\bf Parameters $N$ and $s$ in DASM.} 
% We provide performance comparisons for different combinations of $N$ and $s$  in \cref{tab:re-table3}. The performance is evaluated on the DRealSR test dataset.
% $N$ is set to 4, and $s$ is set to 50 in our setting (bold in the table).
% Larger or smaller $N$ will degrade performance, possibly because it is related to regularization strength. We prefer to use a smaller $N$ because DASM is computationally time-consuming. Therefore, after carefully balancing training duration and model performance, we select $N$=4 as the final value.
% Small $s$ will have similar performance, but the image quality will be compromised when setting large $s$. Our experimental results indicate that selecting $s$ within the range of 25-75 may yield better performance.
We compare performance under different combinations of $N$ and $s$ in \cref{tab:re-table3}. The evaluation is conducted on the DRealSR test dataset. In our setting, $N$ is set to 4 and $s$ to 50 (highlighted in bold in the table). Performance degrades when $N$ is either larger or smaller, possibly due to its effect on regularization strength. Since DASM is computationally expensive, we prefer a smaller $N$. After balancing training time and performance, we select $N=4$ as the final value. Smaller values of $s$ yield similar performance, while larger values degrade image quality. Experimental results suggest that choosing $s$ between 25 and 75 achieves better overall performance.

\begin{table}[t]
\centering
\caption{Ablation studies for hyperparameter $N$ and $s$.}
\label{tab:re-table3}
    \resizebox{0.48\textwidth}{!}{
    \large
    \begin{tabular}{*{10}{c}}
        \toprule
       \textbf{$N$} & \textbf{$s$} &
       \textbf{LPIPS$\downarrow$} &
       \textbf{DISTS $\downarrow$} & 
       \textbf{FID $\downarrow$} & 
       \textbf{MUSIQ $\uparrow$} & 
       \textbf{MANIQA $\uparrow$} &  
       \textbf{CLIPIQA $\uparrow$}  \\
        \midrule
        2 & 50  & 0.3104 & 0.2327 & 137.64 & 64.49 & 0.5717 & 0.7118 \\
        \textbf{4} & \textbf{50}  & \redcolor{0.2967} & \redcolor{0.2136} & \bluecolor{134.98}  & \redcolor{66.62} & \bluecolor{0.5874} & \redcolor{0.7344}\\
        8 & 50  & 0.3421 & 0.2633 & 151.64  & 65.73 & \redcolor{0.5875}& 0.7227 \\
        4 & 25  & \bluecolor{0.3063} & \bluecolor{0.2201} & \redcolor{130.74}  & \bluecolor{66.19} & 0.5828 & \bluecolor{0.7269} \\
        4 & 100  & 0.3176 & 0.2230 & 135.54 &  65.23 & 0.5782  & 0.7024 \\
        \bottomrule
    \end{tabular}
    }
    % \vspace{-0.8em}
\end{table}
\section{Conclusion and Limitation}
\label{sec:conclu}
% We propose TSD-SR, an effective and one-step model based on diffusion prior for Real-ISR. TSD-SR utilizes the TSD to enhance the realism of images generated by the distillation model. And it leverages DASM to sample distribution-based samples and accumulate their gradients to enhance the recovery of details. Our experiments have demonstrated that TSD-SR outperforms existing one-step Real-ISR models in both performance and inference speed.
We propose TSD-SR, an effective one-step model for Real-ISR based on diffusion priors. TSD-SR utilizes TSD to enhance the realism of images generated by the distillation model, and leverages DASM to sample distribution-based noisy latents and accumulate their gradients, thereby improving detail recovery. Our experiments demonstrate that TSD-SR outperforms existing one-step Real-ISR models in both restoration quality and inference speed.

\noindent {\bf limitations.} 
% Although our model boasts excellent inference speed and restoration performance, it still has a large number of model parameters compared to the past GAN/non-diffusion approaches. In the future, we plan to employ pruning or quantization methods to compress the model parameters, striving for a lightweight and efficient Real-ISR model.
Although our model achieves excellent inference speed and restoration performance, it still contains significantly more parameters compared to previous GAN- or non-diffusion-based methods. In future work, we plan to apply pruning or quantization techniques to compress the model, aiming to develop a lightweight and efficient Real-ISR solution.

\section{Acknowledgment}
 This work was supported by National Natural Science Foundation of China (62293554, U2336212), Zhejiang Provincial Natural Science Foundation of China (LD24F020007),  National Key R\&D Program of China (SQ2023AAA01005), Ningbo Innovation ``Yongjiang 2035" Key Research and Development Programme (2024Z292), and Young Elite Scientists Sponsorship Program by CAST (2023QNRC001).
\clearpage
% \setcounter{page}{1}
% \maketitlesupplementary
\section*{Supplementary Material}
In this Supplementary Material, we provide additional details, including the
comparison with GAN-based methods in \cref{ap:gan}, more visual comparisons in \cref{ap:vis}, comparisons of full-reference metrics and human preference in \cref{ap:psnr}, theory of Target Score Matching in \cref{ap:theory} and algorithm in \cref{ap:algorithm}. We conduct these additional comparisons and analyses to validate the effectiveness of TSD-SR.

\appendix
% Comparison with GAN-based Methods--------------
\section{Comparison with GAN-based Methods}
\label{ap:gan}
\begin{figure*}[!htbp]
  \centering
  \begin{subfigure}{0.95\linewidth}
    \includegraphics[width=\linewidth]{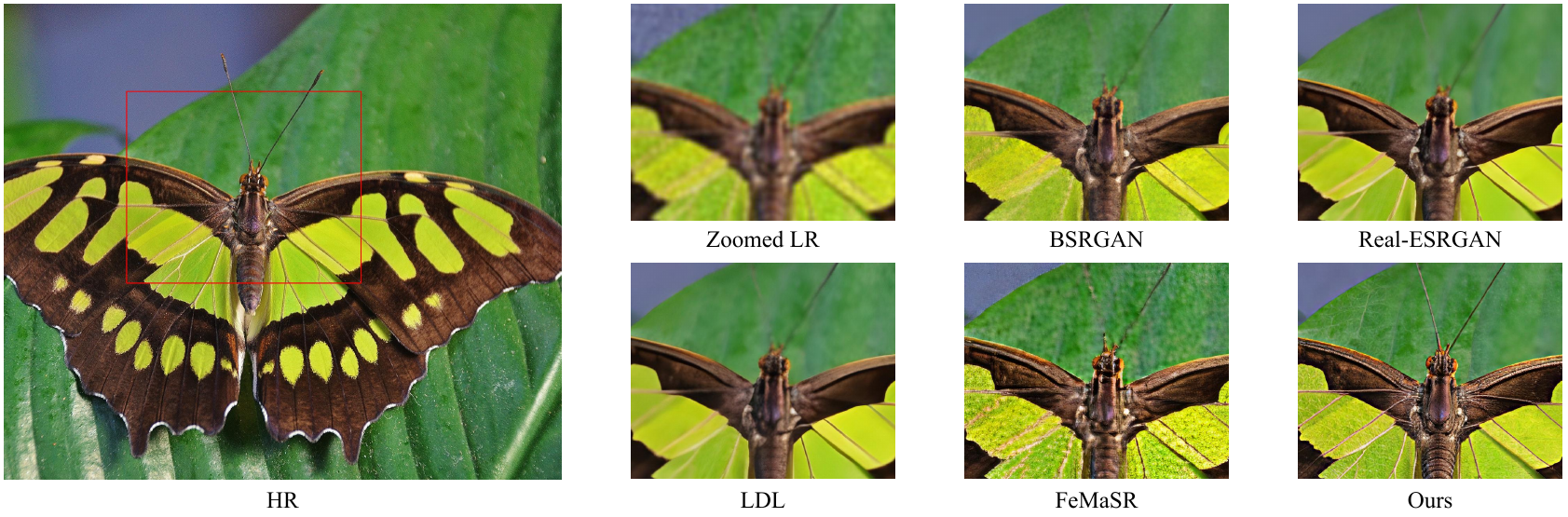}
  \end{subfigure}
\begin{subfigure}{0.95\linewidth}
    \includegraphics[width=\linewidth]{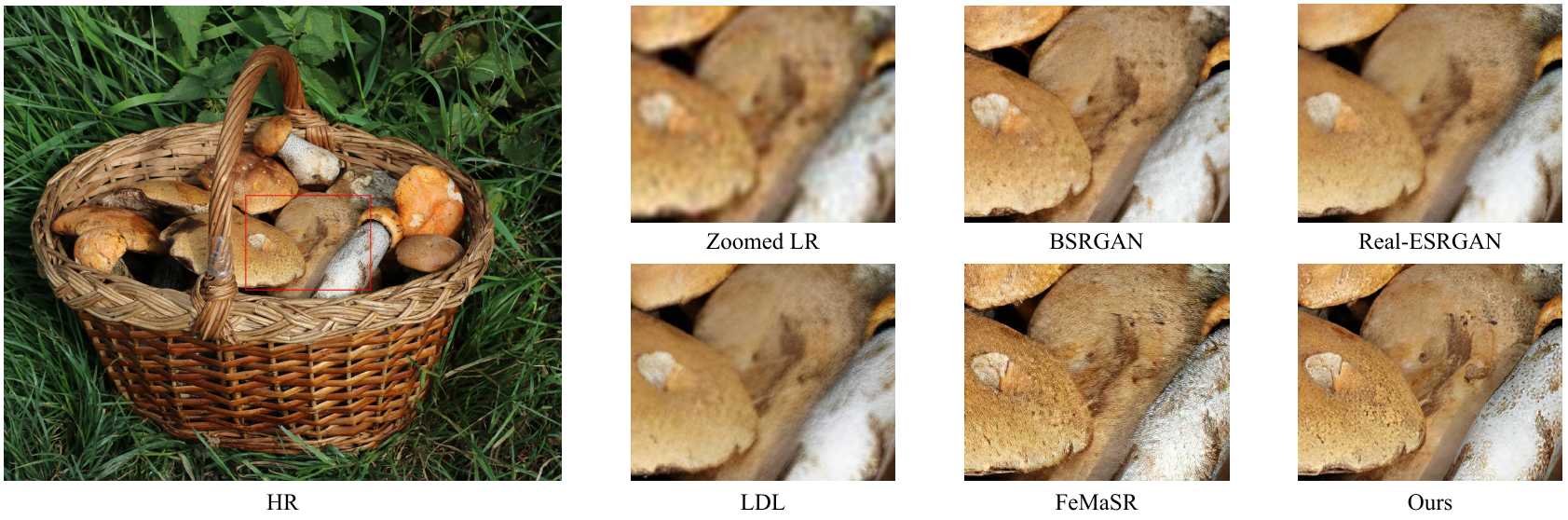}
  \end{subfigure}
\begin{subfigure}{0.95\linewidth}
    \includegraphics[width=\linewidth]{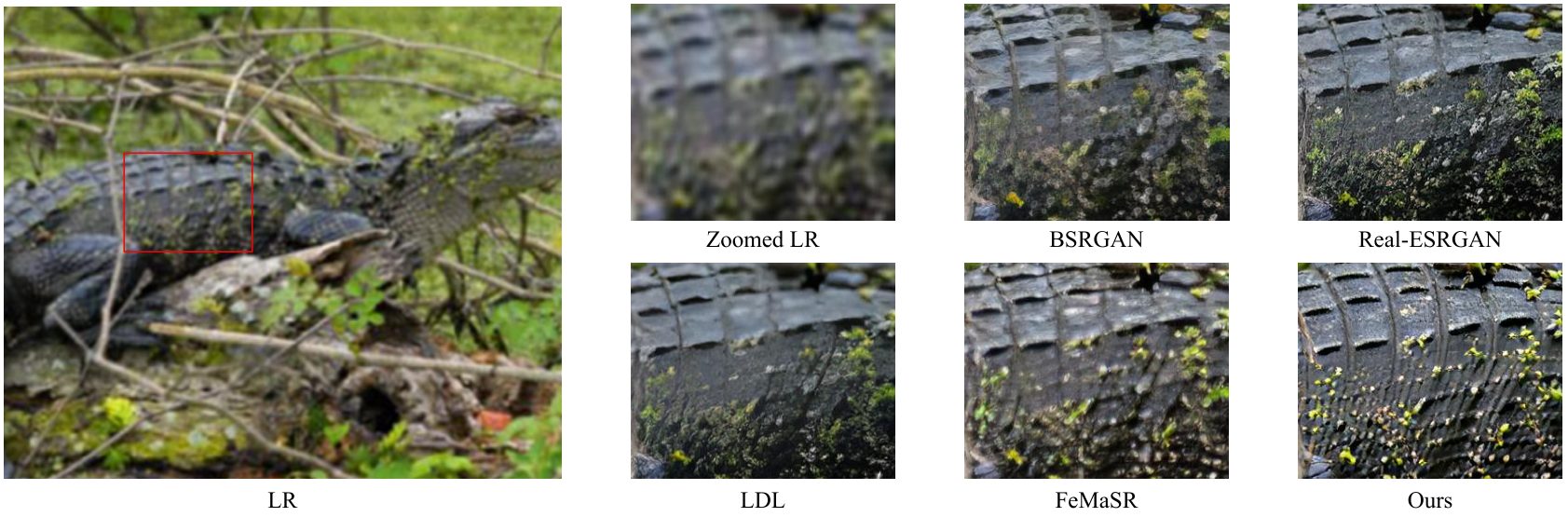}
  \end{subfigure}
  \begin{subfigure}{0.95\linewidth}
    \includegraphics[width=\linewidth]{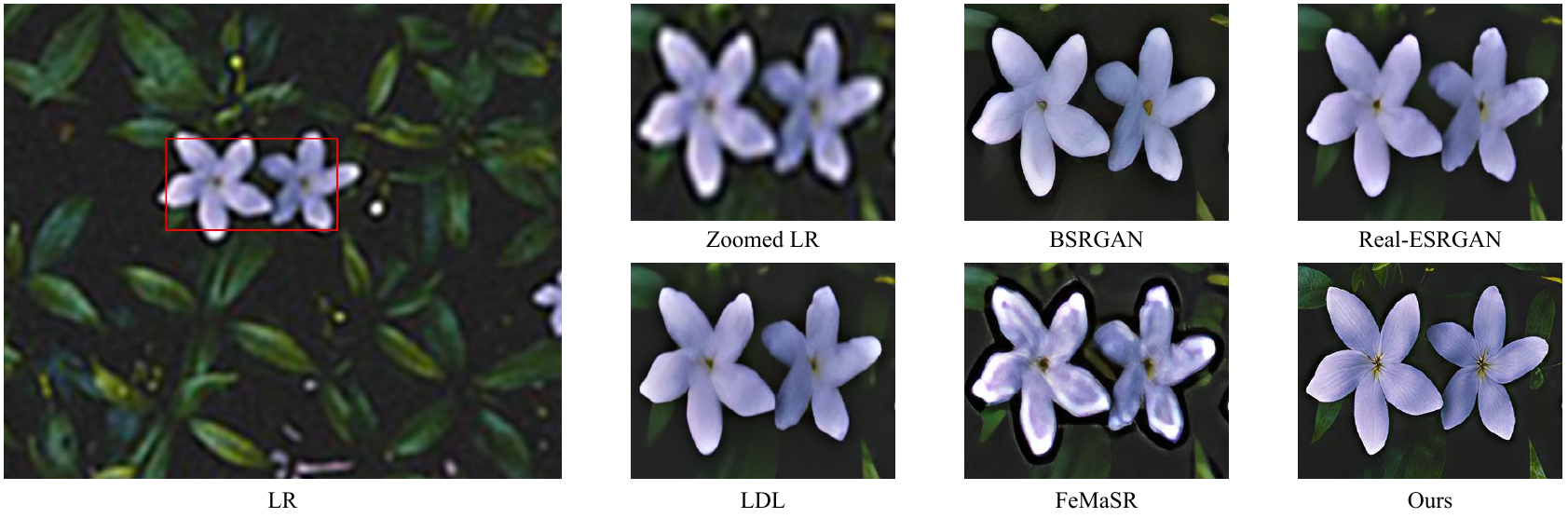}
  \end{subfigure}
   \caption{Qualitative comparisons between TSD-SR and GAN-based Real-ISR methods. Please zoom in for a better view.}
   \label{fig:visualization-gan}
\end{figure*}

We compare our method with GAN-based approaches in \cref{tab:gan}. While the GAN methods show advantages in full-reference metrics such as PSNR and SSIM,  our model outperforms them across all no-reference metrics.
% Some researchers have found the limitations of PSNR and SSIM in the field of image super-resolution \cite{yu2024scaling,xie2024addsr}. 
Prior studies have highlighted the limitations of PSNR and SSIM for evaluating image super-resolution performance \cite{yu2024scaling,xie2024addsr}. 
Their effectiveness in assessing image fidelity in complex degradation scenarios remains debatable, as pixel-level misalignment often arises when restoring severely degraded images.
% However, no-reference metrics assess image quality based on the individual image, without the need to forcibly align with the ground truth. Therefore, in more complex and realistic degradation scenarios, no-reference metrics may be more suitable for evaluating the results of image super-resolution.
However, no-reference metrics evaluate image quality based solely on the individual image, without requiring alignment with the ground truth. Therefore, in more complex and realistic degradation scenarios, they may offer a more appropriate evaluation of super-resolution results.
% In \cref{ap:psnr}, we further provide the visual comparison between full-reference metrics and human preferences, and in \cref{fig:visualization-gan}, we present a visual comparison with GAN-based methods. From the visualization, it can be observed that our model achieves better results in terms of texture details compared to GAN methods.
In \cref{ap:psnr}, we further provide a visual comparison between full-reference metrics and human preferences, and in \cref{fig:visualization-gan}, we present a visual comparison with GAN-based methods. From these visualizations, it is evident that our model produces more realistic texture details than the GAN-based approaches.
\begin{table*}[!b]
\caption{Quantitative comparison with GAN-based methods on both synthetic and real-world benchmarks. The best results of each metric are highlighted in \textcolor{red}{red}.}
\label{tab:gan}
\resizebox{\textwidth}{!}{%
\begin{tabular}{@{}c|c|ccccccccc@{}}
\toprule
\textbf{Datasets} & \textbf{Method} & \textbf{PSNR $\uparrow$} & \textbf{SSIM $\uparrow$} & \textbf{LPIPS $\downarrow$ } & \textbf{DISTS $\downarrow$} & \textbf{FID $\downarrow$} & \textbf{NIQE $\downarrow$} & \textbf{MUSIQ $\uparrow$} & \textbf{MANIQA $\uparrow$} & \textbf{CLIPIQA $\uparrow$} \\ \midrule
 & BSRGAN & {\color[HTML]{D83931} \textbf{28.70}} & 0.8028 & 0.2858 & 0.2143 & 155.61 & 6.5408 & 57.15 & 0.4847 & 0.5091 \\
 & Real-ESRGAN & 28.61 & 0.8051 & 0.2818 & {\color[HTML]{D83931} \textbf{0.2088}} & 147.66 & 6.7001 & 54.27 & 0.4888 & 0.4521 \\
 & LDL & 28.20 & {\color[HTML]{D83931} \textbf{0.8124}} & {\color[HTML]{D83931} \textbf{0.2791}} & 0.2127 & 155.51 & 7.1448 & 53.94 & 0.4894 & 0.4476 \\
 & FeMASR & 26.87 & 0.7569 & 0.3156 & 0.2238 & 157.72 & 5.9067 & 53.70 & 0.4413 & 0.5633 \\
\multirow{-5}{*}{\textbf{DRealSR}} & Ours & 27.77 & 0.7559 & 0.2967 & 0.2136 & {\color[HTML]{D83931} \textbf{134.98}} & {\color[HTML]{D83931} \textbf{5.9131}} & {\color[HTML]{D83931} \textbf{66.62}} & {\color[HTML]{D83931} \textbf{0.5874}} & {\color[HTML]{D83931} \textbf{0.7344}} \\ \midrule
 & BSRGAN & 26.38 & {\color[HTML]{D83931} \textbf{0.7651}} & {\color[HTML]{D83931} \textbf{0.2656}} & 0.2123 & 141.24 & 5.6431 & 63.28 & 0.5425 & 0.5114 \\
 & Real-ESRGAN & {\color[HTML]{D83931} \textbf{26.65}} & 0.7603 & 0.2726 & {\color[HTML]{D83931} \textbf{0.2065}} & 136.29 & 5.8471 & 60.45 & 0.5507 & 0.4518 \\
 & LDL & 25.28 & 0.7565 & 0.2750 & 0.2119 & 142.74 & 5.9880 & 60.92 & 0.5494 & 0.4559 \\
 & FeMASR & 25.06 & 0.7356 & 0.2936 & 0.2285 & 141.01 & 5.7696 & 59.05 & 0.4872 & 0.5405 \\
\multirow{-5}{*}{\textbf{RealSR}} & Ours & 24.81 & 0.7172 & 0.2743 & 0.2104 & {\color[HTML]{D83931} \textbf{114.45}} & {\color[HTML]{D83931} \textbf{5.1298}} & {\color[HTML]{D83931} \textbf{71.19}} & {\color[HTML]{D83931} \textbf{0.6347}} & {\color[HTML]{D83931} \textbf{0.7160}} \\ \midrule
 & BSRGAN & {\color[HTML]{D83931} \textbf{24.58}} & 0.6269 & 0.3502 & 0.2280 & 49.55 & 4.7501 & 61.68 & 0.4979 & 0.5386 \\
 & Real-ESRGAN & 24.02 & {\color[HTML]{D83931} \textbf{0.6387}} & 0.3150 & 0.2123 & 38.87 & 4.8271 & 60.38 & 0.5401 & 0.5251 \\
 & LDL & 23.83 & 0.6344 & 0.3256 & 0.2227 & 42.28 & 4.8555 & 60.04 & 0.5328 & 0.5180 \\
 & FeMASR & 22.45 & 0.5858 & 0.3370 & 0.2205 & 41.97 & 4.8679 & 57.94 & 0.4787 & 0.5769 \\
\multirow{-5}{*}{\textbf{DIV2K-Val}} & Ours & 23.02 & 0.5808 & {\color[HTML]{D83931} \textbf{0.2673}} & {\color[HTML]{D83931} \textbf{0.1821}} & {\color[HTML]{D83931} \textbf{29.16}} & {\color[HTML]{D83931} \textbf{4.3244}} & {\color[HTML]{D83931} \textbf{71.69}} & {\color[HTML]{D83931} \textbf{0.6192}} & {\color[HTML]{D83931} \textbf{0.7416}} \\ \bottomrule
\end{tabular}%
}
\end{table*}

% More Visual Comparisons ---------------------------
\section{More Visual Comparisons}
\label{ap:vis}
\begin{figure*}[!b]
  \centering
  \begin{subfigure}{\linewidth}
    \includegraphics[width=\linewidth]{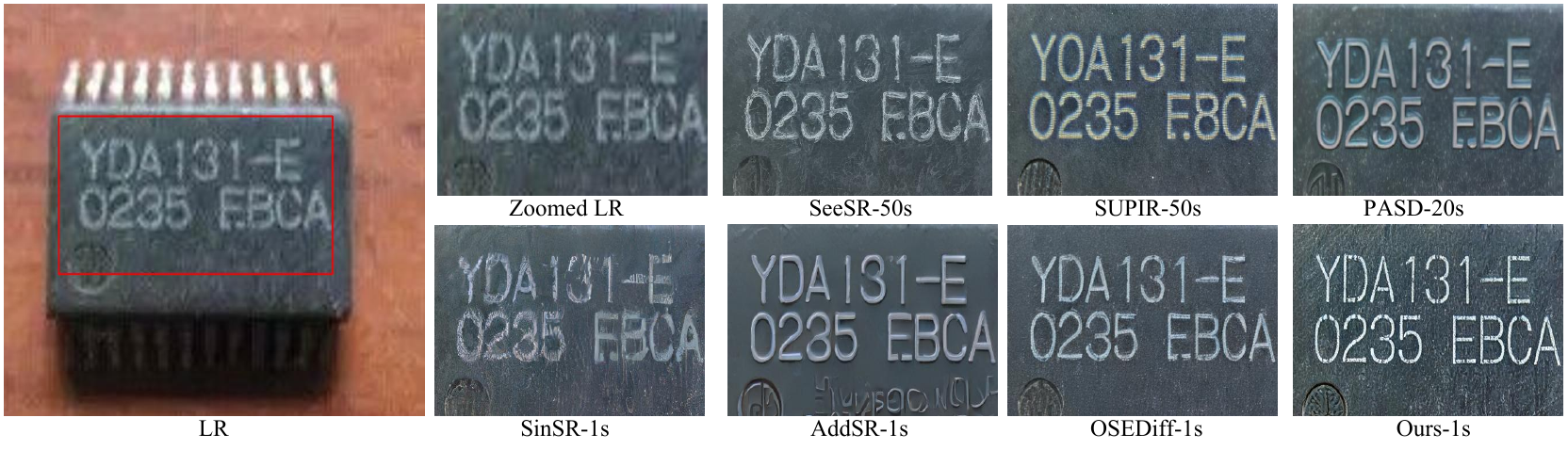}
  \end{subfigure}
     \caption{Qualitative comparisons between TSD-SR and different diffusion-based methods. Our method can effectively restore the texture and details of the corresponding object under challenging degradation conditions. Please zoom in for a better view.}
     \label{fig:visualization4}
\end{figure*}

\begin{figure*}[!htbp]
  \centering
  \begin{subfigure}{\linewidth}
    \includegraphics[width=\linewidth]{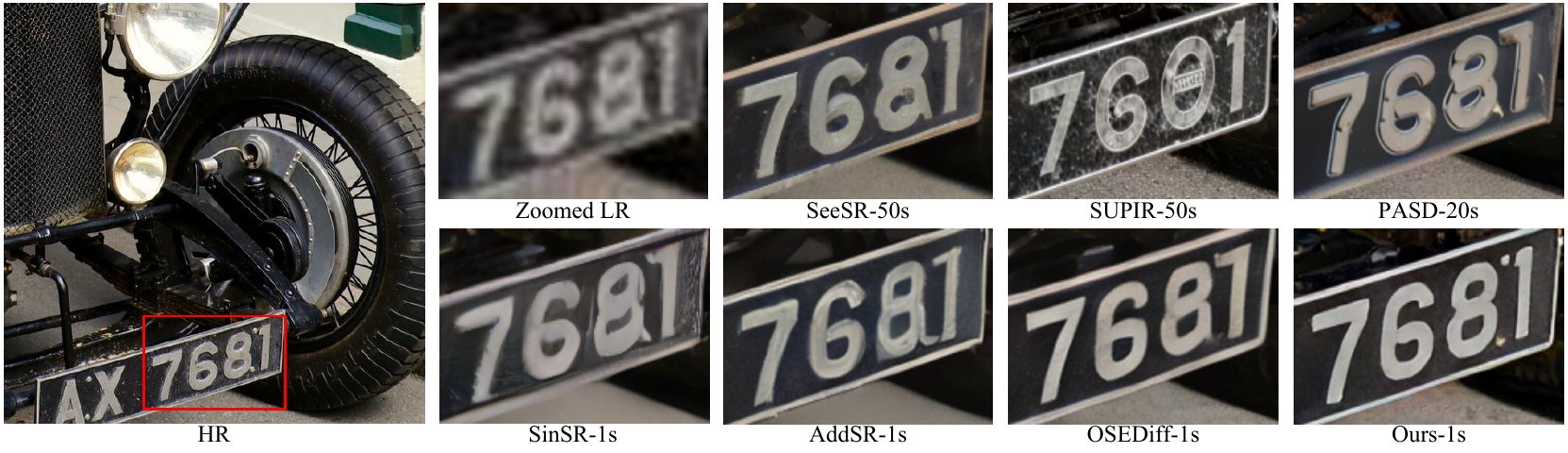}
  \end{subfigure}
 \begin{subfigure}{\linewidth}
    \includegraphics[width=\linewidth]{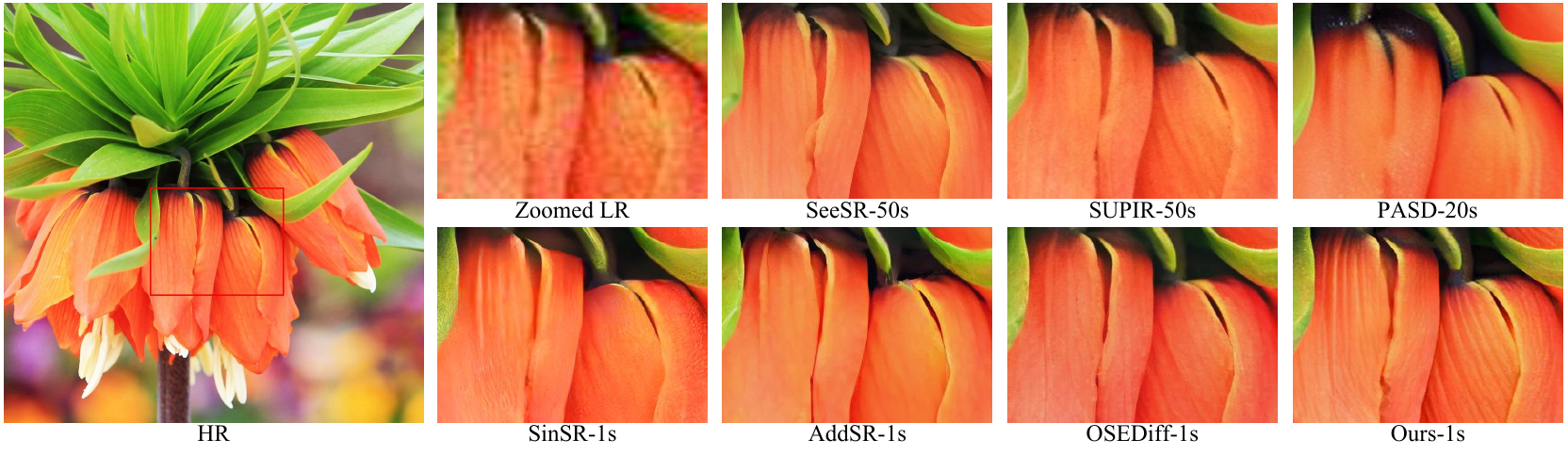}
  \end{subfigure}
    \begin{subfigure}{\linewidth}
    \includegraphics[width=\linewidth]{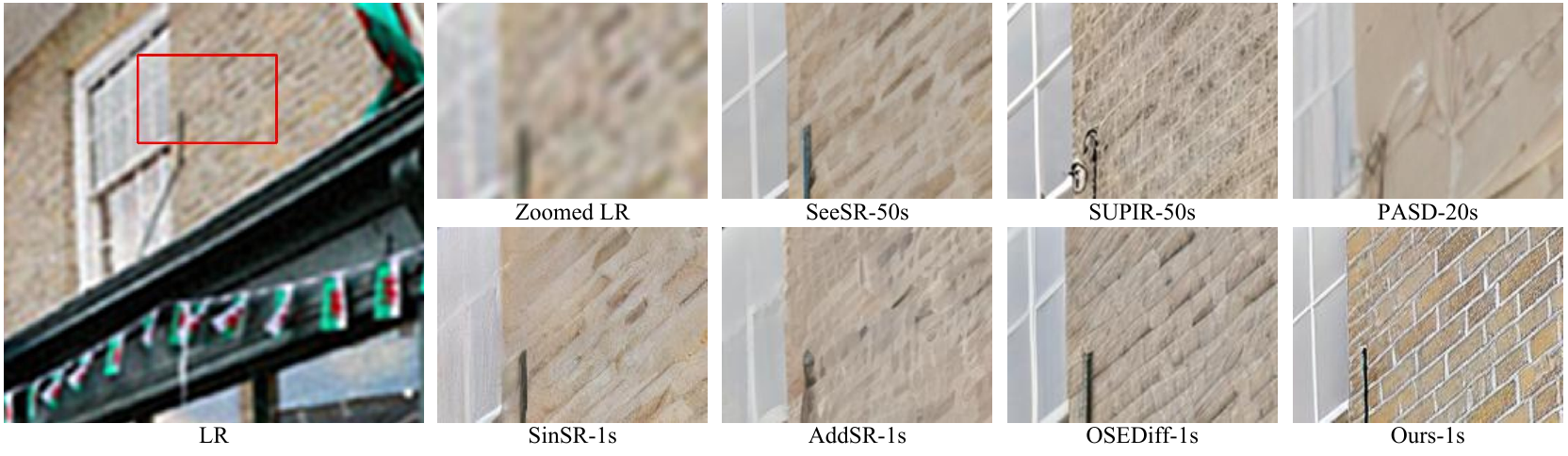}
  \end{subfigure}
      \begin{subfigure}{\linewidth}
    \includegraphics[width=\linewidth]{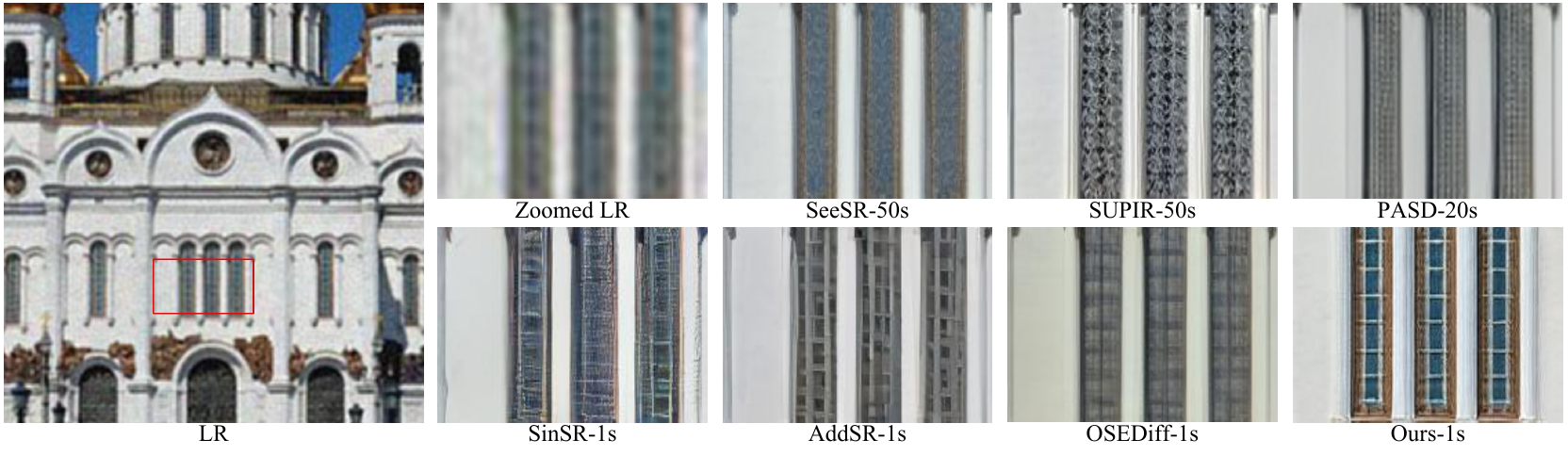}
  \end{subfigure}
   \caption{Qualitative comparisons between TSD-SR and different diffusion-based methods. Our method can effectively restore the texture and details of the corresponding object under challenging degradation conditions. Please zoom in for a better view.}
   \label{fig:visualization5}
\end{figure*}

\begin{figure*}[!htbp]
  \centering
  \begin{subfigure}{\linewidth}
    \includegraphics[width=\linewidth]{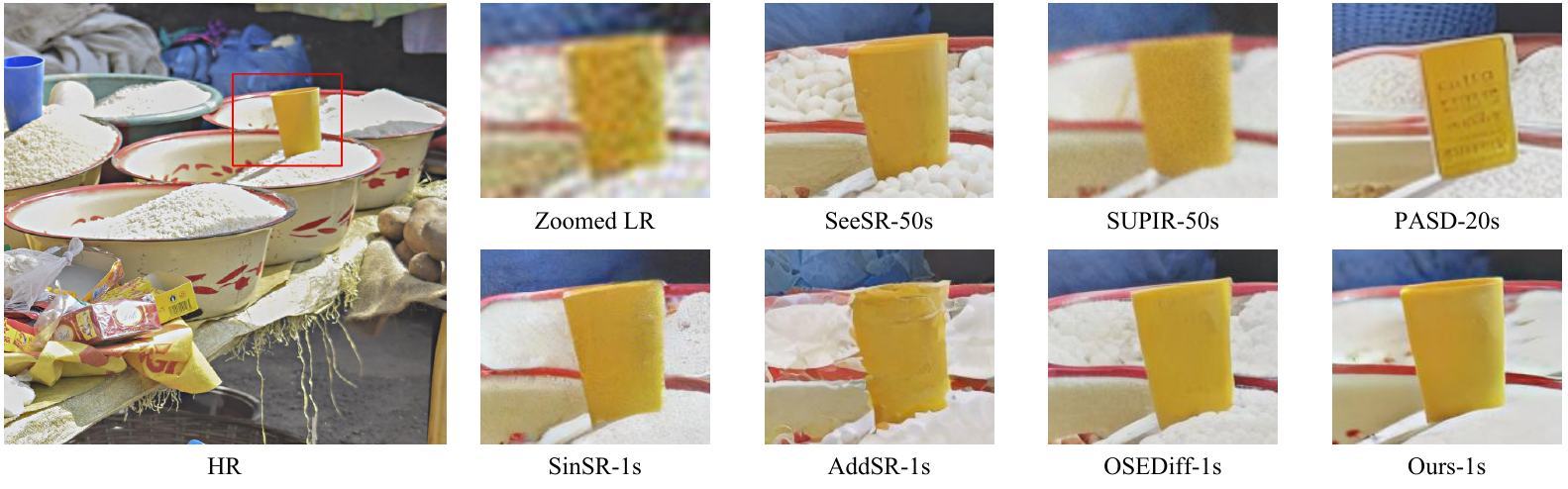}
  \end{subfigure}
    \begin{subfigure}{\linewidth}
    \includegraphics[width=\linewidth]{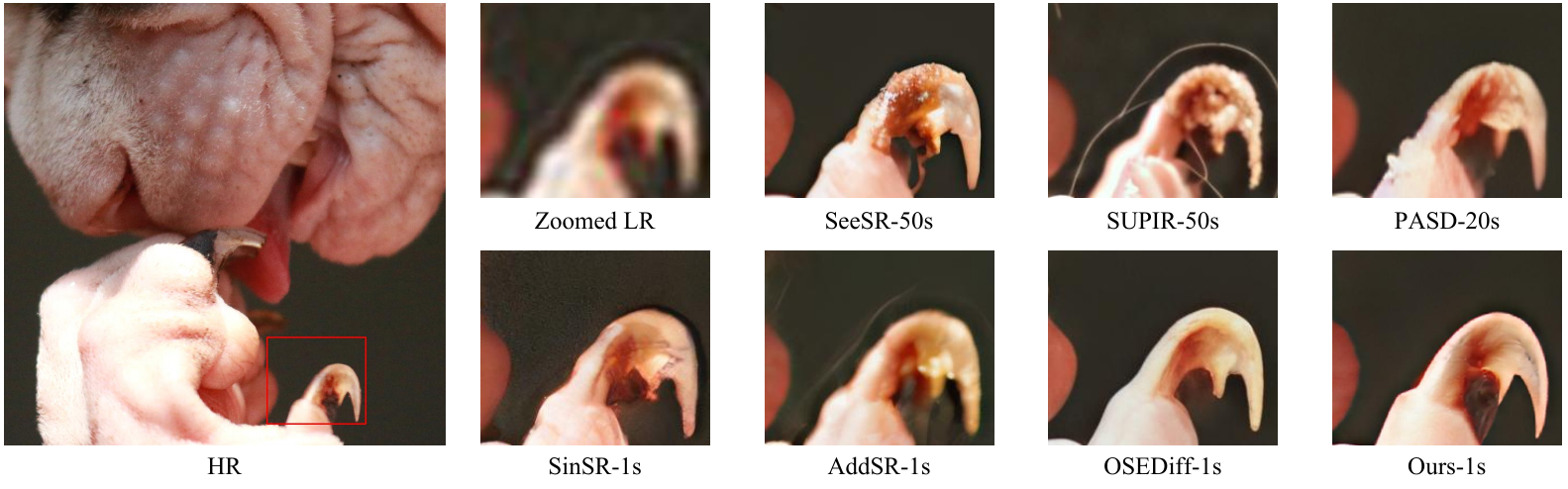}
  \end{subfigure}
      \begin{subfigure}{\linewidth}
    \includegraphics[width=\linewidth]{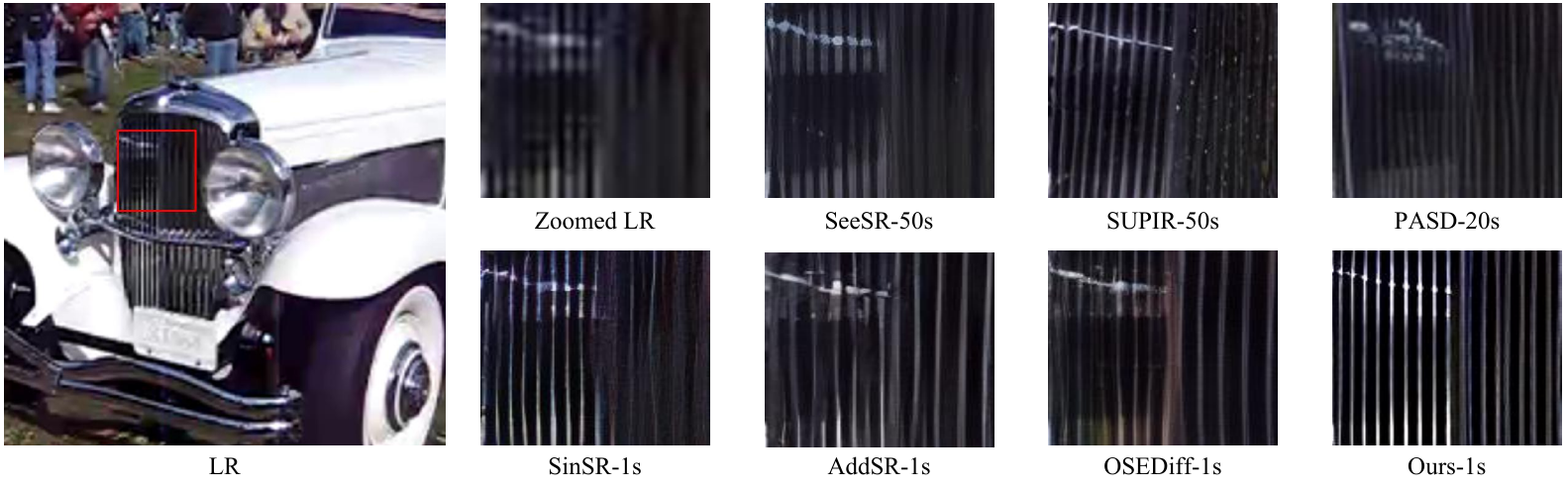}
  \end{subfigure}
        \begin{subfigure}{\linewidth}
    \includegraphics[width=\linewidth]{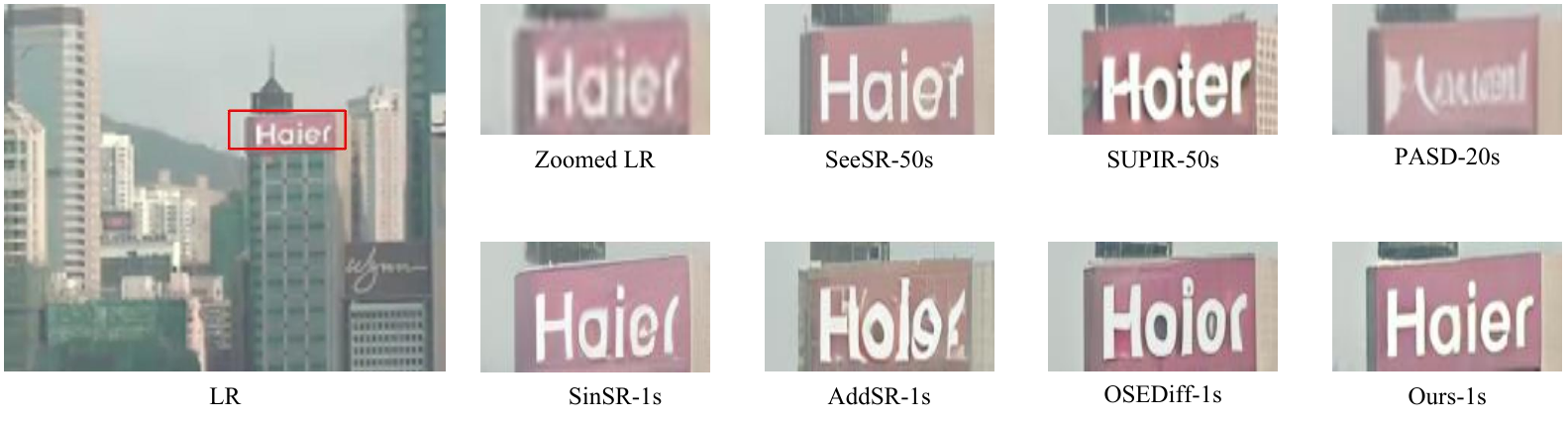}
  \end{subfigure}
   \caption{Qualitative comparisons between TSD-SR and different diffusion-based methods. Our method can effectively restore the texture and details of the corresponding object under challenging degradation conditions. Please zoom in for a better view.}
   \label{fig:visualization6}
\end{figure*}

% In \cref{fig:visualization4,fig:visualization5,fig:visualization6}, we provide more visual comparisons with other diffusion-based methods. Numerous examples demonstrate the robust restoration capabilities of TSD-SR and the high quality of restored images. 
In \cref{fig:visualization4,fig:visualization5,fig:visualization6}, we provide additional visual comparisons with other diffusion-based methods. These examples further demonstrate the robust restoration capabilities of TSD-SR and the high quality of the restored images.

\section{Comparisons of Full-reference Metrics and Human Preference}
\label{ap:psnr}
\begin{figure*}[!t]
  \centering
  \begin{subfigure}{0.48\linewidth}
    \includegraphics[width=\linewidth, height=0.65\linewidth]{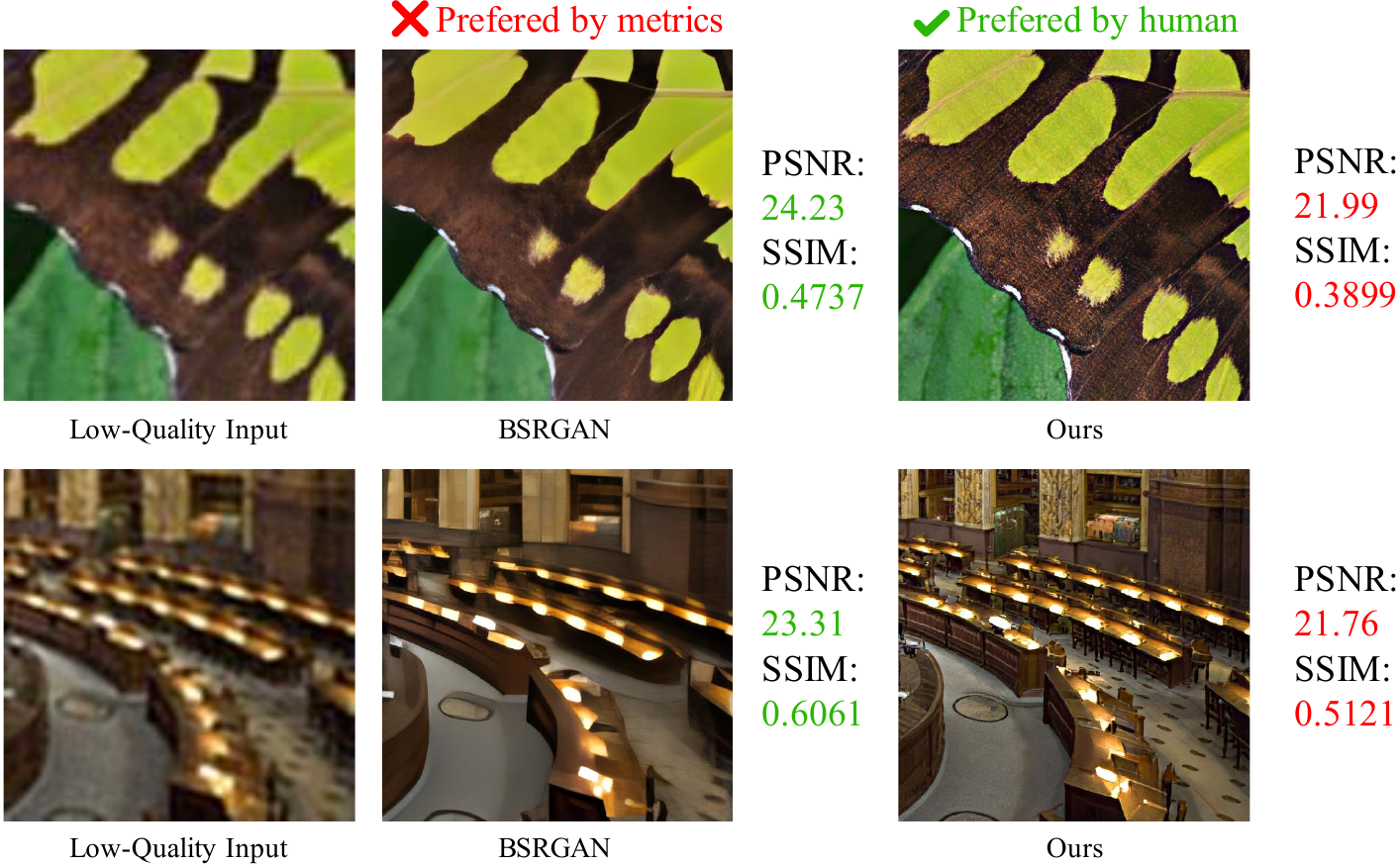}
  \end{subfigure}
  \hfill
  \begin{subfigure}{0.48\linewidth }
    \includegraphics[width=\linewidth, height=0.65\linewidth]{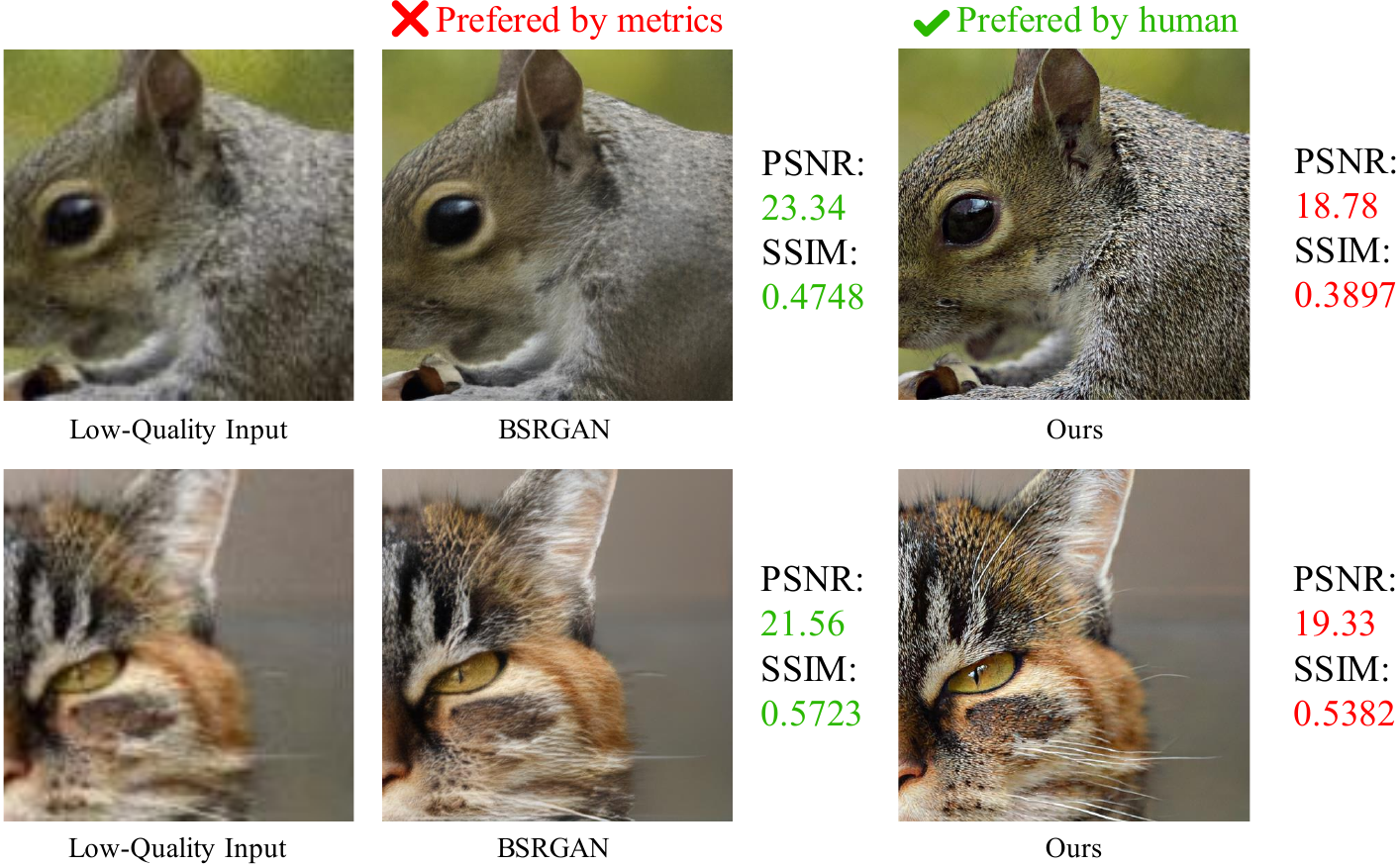}
  \end{subfigure}
\begin{subfigure}{0.48\linewidth }
    \includegraphics[width=\linewidth, height=0.65\linewidth]{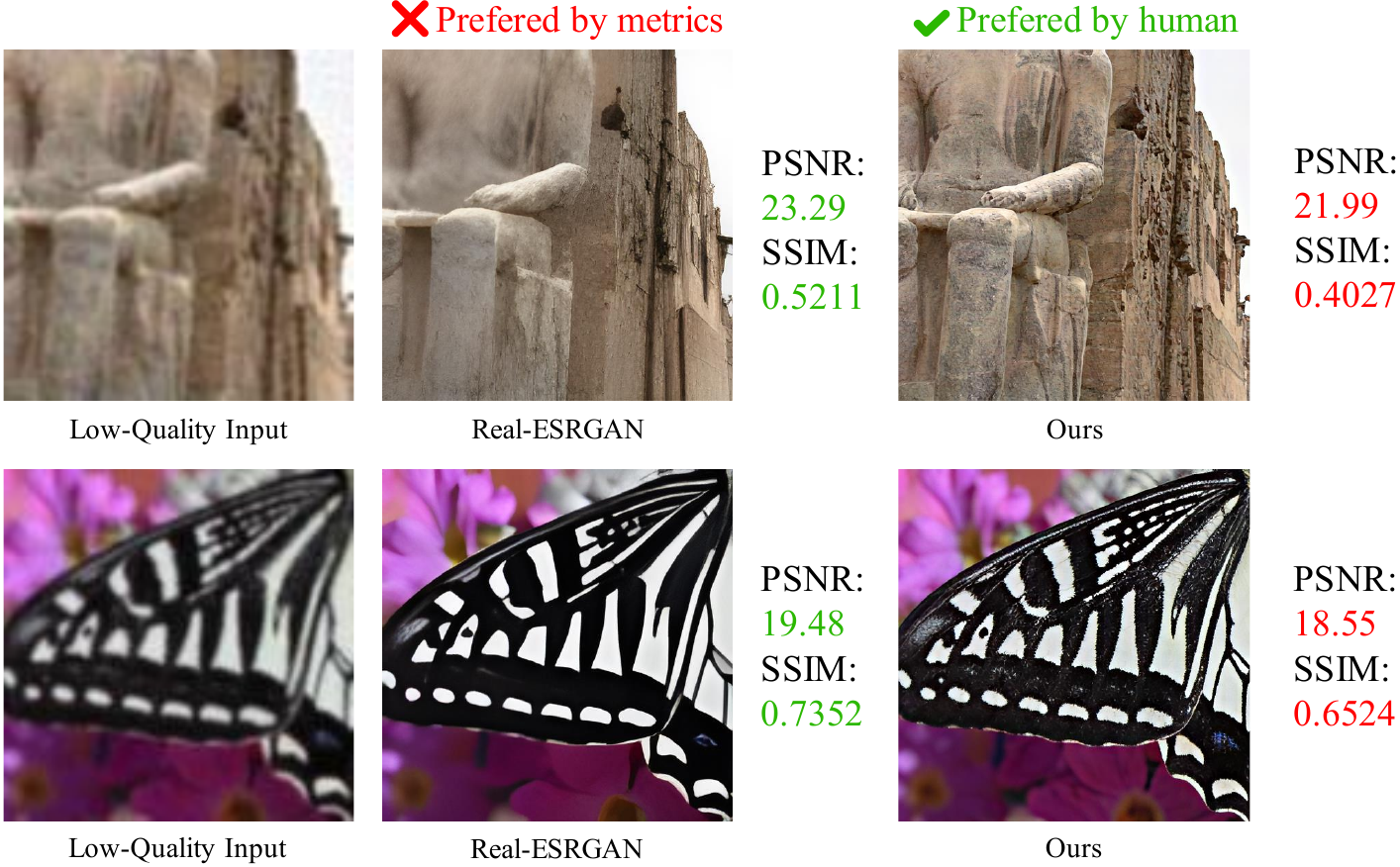}
  \end{subfigure}
  \hfill
    \begin{subfigure}{0.48\linewidth }
    \includegraphics[width=\linewidth, height=0.65\linewidth]{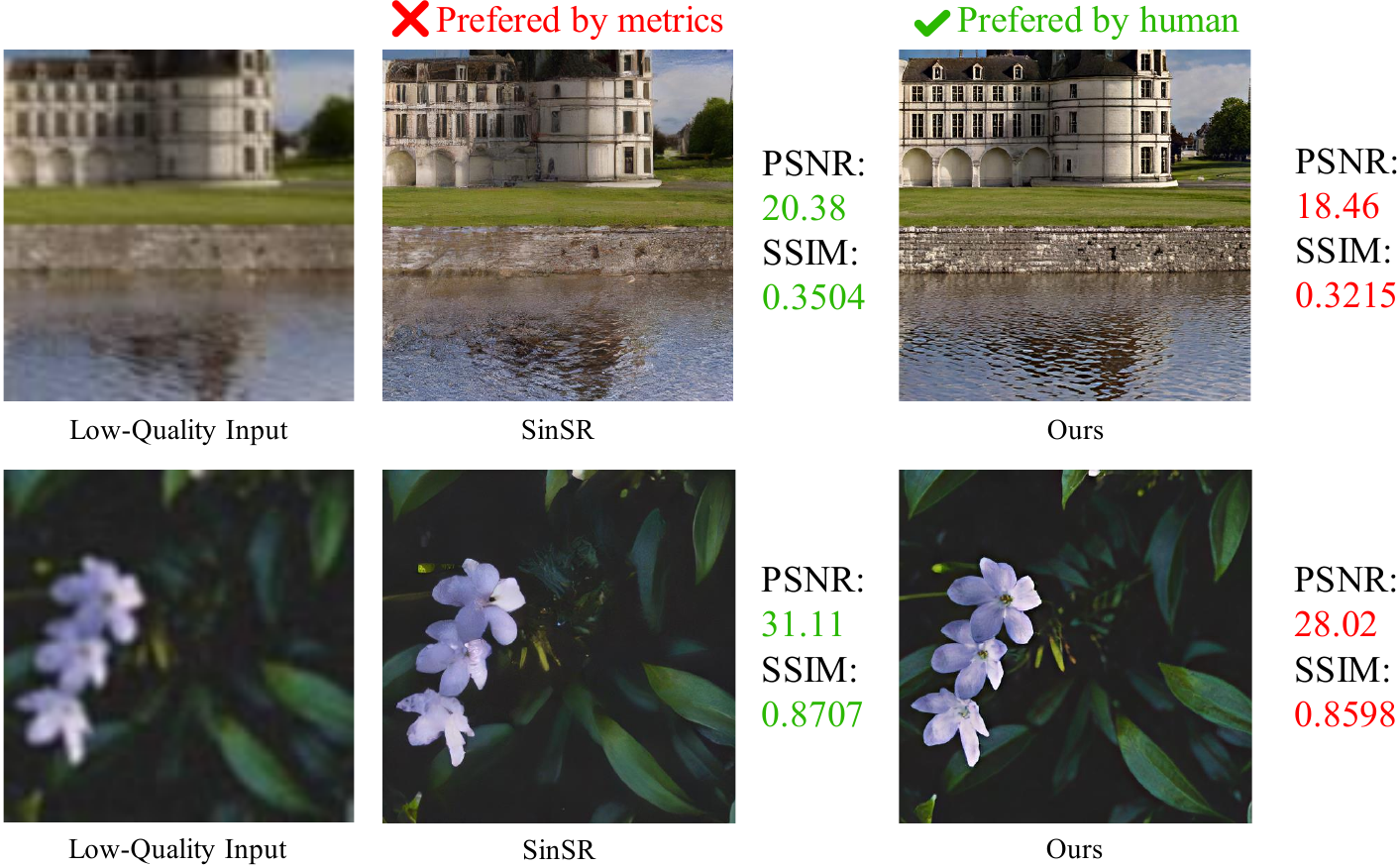}
  \end{subfigure}
    \begin{subfigure}{0.48\linewidth }
    \includegraphics[width=\linewidth, height=0.65\linewidth]{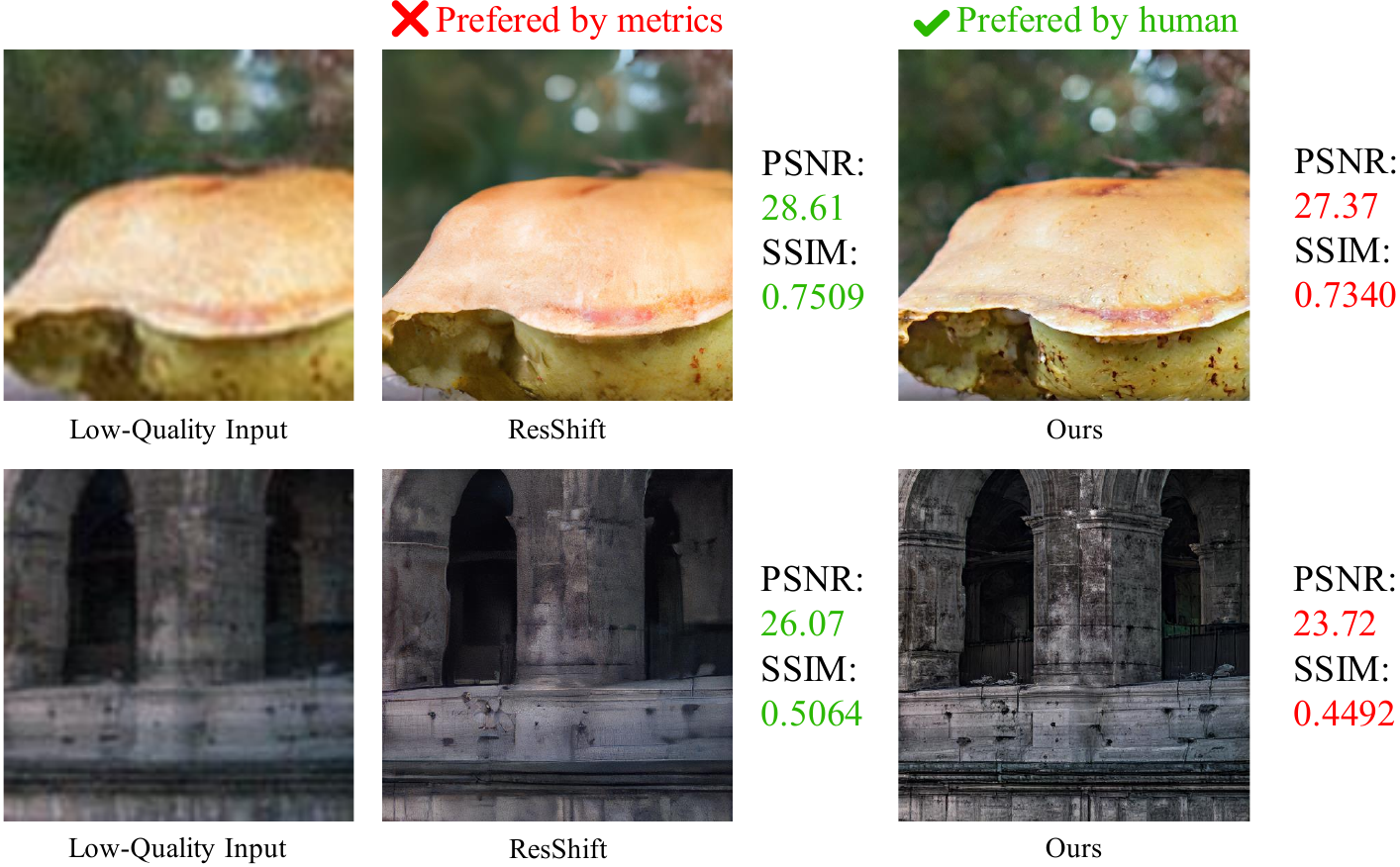}
  \end{subfigure}
  \hfill
    \begin{subfigure}{0.48\linewidth }
    \includegraphics[width=\linewidth, height=0.65\linewidth]{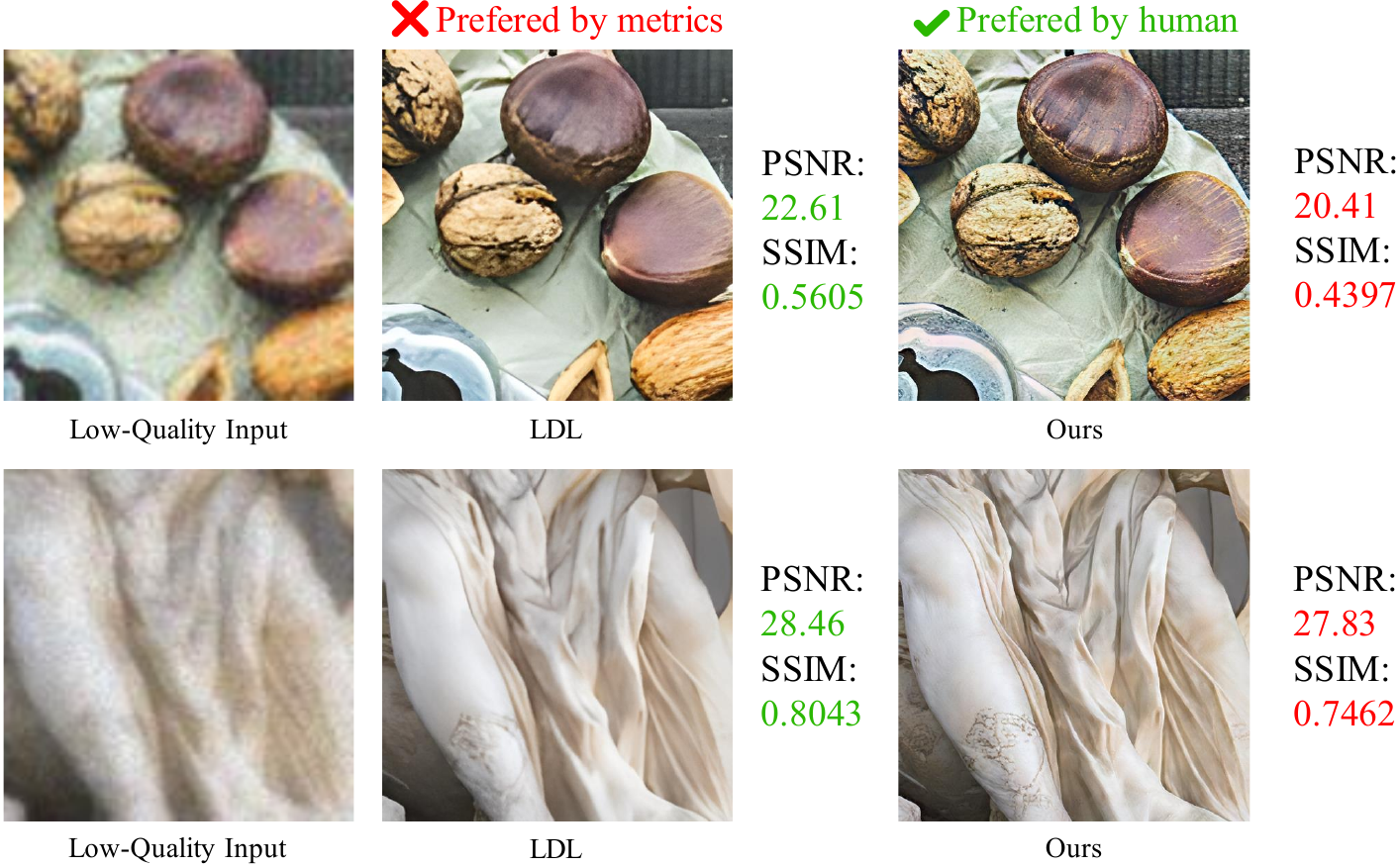}
  \end{subfigure}
  \caption{Comparisons between full-reference metric assessments and human visual preference. Despite scoring lower on full-reference metrics, TSD-SR generates images that align with human preference.}
    \label{apfig:metric}
\end{figure*}

% We present additional comparative experiments in \Cref{apfig:metric} to demonstrate that PSNR and SSIM may have limitations in assessing image fidelity under complex degradation scenarios. It can be observed that GAN-based methods with higher PSNR and SSIM produce over-smooth or broken textures, raising concerns about their realism and fidelity. While our approach trades off PSNR and SSIM for natural detail restoration, it achieves enhanced realism and broader perceptual acceptance (Our additional user study reveals that 90.28\% of participants prefer ours instead of high PSNR and SSIM methods.).
We present additional comparative experiments in \Cref{apfig:metric} to demonstrate the limitations of PSNR and SSIM in assessing image fidelity under complex degradation scenarios. As observed, GAN-based methods with higher PSNR and SSIM scores tend to produce over-smoothed or fragmented textures, raising concerns about their realism and perceptual fidelity. In contrast, our approach sacrifices some PSNR and SSIM performance to achieve more natural detail restoration, resulting in enhanced realism and broader perceptual acceptance. An additional user study shows that 90.28\% of participants prefer our results over those of methods with higher PSNR and SSIM scores.
% PSNR Compare ---------------------------
% Theory of Target Score Distillation---------
\section{Theory of Target Score Matching}
\label{ap:theory}
The core idea of Target Score Matching (TSM) is that for samples drawn from the same distribution, the real scores predicted by the Teacher Model should be close to each other. Thus, we minimize the MSE loss between the Teacher Model’s predictions on $\boldsymbol{\hat z}_t$ and $\boldsymbol{ z}_t$ by
\begin{equation}
\begin{split}
% & \nabla_\theta\mathcal{L}_{\mathrm{TSM}}(\boldsymbol{\hat z },\boldsymbol{z},c_y) \\ 
% & = \mathbb{E}_{t,\epsilon}\left[w(t)({\epsilon}_\psi(\boldsymbol{\hat z}_t;t,c_y)-{\epsilon}_\psi(\boldsymbol{z}_t;t,c_y))\frac{\partial\boldsymbol{\hat z}}{\partial\theta}\right]
& \mathcal{L}_{\mathrm{MSE}}(\boldsymbol{\hat z},\boldsymbol{ z},c_y) \\
& =\mathbb{E}_{t,\epsilon}\left[w(t)\|\epsilon_{\psi}(\boldsymbol{\hat z}_t;t,c_y)-\epsilon_{\psi}( \boldsymbol{z}_t;t,c_y)\|_{2}^{2}\right]
\label{apeq:mse}
\end{split}
\end{equation}
where the expectation of the gradient is computed across all diffusion timesteps $t \in \{1,\cdots,T\}$ and $\epsilon \sim \mathcal{N}(0,I)$.

To understand the difficulties of this approach, consider the gradient of 
\begin{equation}
\begin{split}
& \nabla_\theta\mathcal{L}_{\mathrm{MSE}}(\boldsymbol{\hat z},\boldsymbol{ z},c_y) 
 =\mathbb{E}_{t,\epsilon} 
 \Big[w(t) \cdot
 \underbrace{\frac{\partial{\epsilon}_\psi(\boldsymbol{\hat z}_t;t,c_y)}{\partial\boldsymbol{\hat z}_t}}_{\text{Diffusion Jacobian}} \\
 & \underbrace{({\epsilon}_\psi(\boldsymbol{\hat z }_t;t,c_y)-{\epsilon}_\psi(\boldsymbol{z}_t;t,c_y))}_{\text{Prediction Residual}} 
  \underbrace{\frac{\partial\boldsymbol{\hat z}}{\partial\theta}}_{\text{Generator Jacobian}}
 \Big]
\label{apeq:msetsm}
\end{split}
\end{equation}
where we absorb $\frac{\partial\boldsymbol{\boldsymbol{\hat z}}_t}{\partial\boldsymbol{\boldsymbol{\hat z}}}$ and the other constant into $w(t)$. The computation of the Diffusion Jacobian term is computationally demanding, as it necessitates backpropagation through the Teacher Model. DreamFusion \cite{poole2022dreamfusion} found that this term struggles with small noise levels due to its training to approximate the scaled Hessian of marginal density. This work also demonstrated that omitting the Diffusion Jacobian term leads to an effective gradient for optimizing. Similar to their approach, we update \cref{apeq:msetsm} by omitting Diffusion Jacobian:
\begin{equation}
\begin{split}
& \nabla_\theta\mathcal{L}_{\mathrm{TSM}}(\boldsymbol{\hat z},\boldsymbol{ z},c_y)= \\
& \mathbb{E}_{t,\epsilon}\Big[w(t)\underbrace{({\epsilon}_\psi(\boldsymbol{\hat  z}_t;c_y,t)-{\epsilon}_\psi(\boldsymbol{z}_t;c_y,t))}_{\text{Prediction Residual}}\underbrace{\frac{\partial\boldsymbol{\hat z}}{\partial\theta}}_{\text{Generator Jacobian}}\Big]
\label{apeq:tsm}
\end{split}
\end{equation}
The effectiveness of the method can be proven by starting from the KL divergence. We can use a Sticking-the-Landing \cite{roeder2017sticking} style gradient by thinking of ${\epsilon}_\psi(\boldsymbol{z}_t;c_y,t)$ as a control variate for $\hat \epsilon$. For detailed proof, refer to Appendix 4 of DreamFusion \cite{poole2022dreamfusion}. It demonstrates that the gradient of this loss yields the same updates as optimizing the training loss $\mathcal{L}_{\mathrm{MSE}}$ \cref{apeq:mse}, excluding the Diffusion Jacobian term.

Compared with the VSD loss, we find that the term ``Prediction Residual" has changed, and the two losses are similar in the gradient update mode. Specifically, we find that VSD employs identical inputs for both the Teacher and LoRA models to compute the gradient, while here TSM uses high-quality and suboptimal inputs for the Teacher Model. The losses are related to each other through $\epsilon_\phi(\boldsymbol{\hat z}_t;t,c_y)$.
\section{Algorithm}
\label{ap:algorithm}
\cref{ap:algo} details our TSD-SR training procedure. We use classifier-free guidance (cfg) for the Teacher Model and the LoRA Model. The cfg weight is set to 7.5.
\begin{algorithm*}[!t]
\caption{TSD-SR Training Procedure}
\SetKwInput{KwInput}{Input}                % Set the Input
\SetKwInput{KwOutput}{Output}              % set the Output
\DontPrintSemicolon
\label{ap:algo}
  \KwInput{$\mathcal{D}$ = $\{x_L,x_H,c_y\}$ , pre-trained Teacher Diffusion Model including VAE encoder $E_\psi$, denoising network $\epsilon_\psi$ and VAE decoder $D_\psi$, the number of iterations $N$ and step size $s$ of DASM.}
  \KwOutput{Trained one-step Student Model $G_\theta$.}
  % \KwData{Testing set $x$}
    {Initialize Student Model $G_\theta$, including 
            $E_\theta \leftarrow E_\psi $ with trainable LoRA,  
            $\epsilon_\theta \leftarrow \epsilon_\psi $ with trainable LoRA,
            $D_\theta \leftarrow D_\psi $.}
            
     Initialize LoRA diffusion network $\epsilon_\phi \leftarrow \epsilon_\psi $ with trainable LoRA.

\While{train}{
    Sample $(x_L,x_H,c_y) \sim \mathcal{D}$
    
    \tcc{Network forward} 
    
    $\boldsymbol{\hat z} \leftarrow \epsilon_\theta(E_\theta(x_L))$, $\boldsymbol{z} \leftarrow E_\psi(x_H)$
    
    $\hat x_H \leftarrow D_\psi(\boldsymbol{\hat z})$

    \tcc{Compute reconstruction loss}

    $\mathcal{L}_{Rec} \leftarrow LPIPS(\hat x_H, x_H)$

    \tcc{Compute regularization loss}

    Sample $\epsilon$ from $\mathcal{N}(0,I)$, $t$ from $\{50,\cdots,950\}$

    $\sigma_t \leftarrow $ FlowMatchingScheduler($t$)
    
    $\boldsymbol{\hat z}_t \leftarrow \sigma_t \epsilon + (1 - \sigma_t) \boldsymbol{\hat z}$, $\boldsymbol{z}_t \leftarrow \sigma_t \epsilon + (1 - \sigma_t) \boldsymbol{z}$

    $\mathcal{L}_{Reg} \leftarrow \mathcal{L}_{TSD}(\boldsymbol{\hat z}_t,\boldsymbol{z}_t,c_y)$ // \cref{eq:tsd}

    \For{$i\leftarrow 1$ \KwTo $N$}{
        $cur \leftarrow t - i \cdot s $

        $pre \leftarrow t - i \cdot s + s $
        
        $\sigma_{cur} \leftarrow $ FlowMatchingScheduler($cur$)
        
        $\sigma_{pre} \leftarrow $ FlowMatchingScheduler($pre$)
        
        $\boldsymbol{\hat z}_{cur} \leftarrow \boldsymbol{\hat z}_{pre} + (\sigma_{cur} - \sigma_{pre}) \cdot \epsilon_\phi(\boldsymbol{\hat z}_{pre};pre,c_y)$ // \cref{eq:sample}

        $\boldsymbol{z}_{cur} \leftarrow \boldsymbol{z}_{pre} + (\sigma_{cur} - \sigma_{pre}) \cdot \epsilon_\psi(\boldsymbol{z}_{pre};pre,c_y)$

        $\mathcal{L}_{Reg}$ +=  $weight \cdot \mathcal{L}_{TSD}(\boldsymbol{\hat z}_{cur},\boldsymbol{z}_{cur},c_y)$
        
    }

    $\mathcal{L}_G \leftarrow \mathcal{L}_{Rec} + \gamma \mathcal{L}_{Reg}$

    Update $\theta$ with $\mathcal{L}_G$
    
    \tcc{Compute diffusion loss for LoRA Model}

    Sample $\epsilon$ from $\mathcal{N}(0,I)$, $t$ from $\{50,\cdots,950\}$

    $\sigma_t \leftarrow $ FlowMatchingScheduler($t$)

    $\boldsymbol{\hat z}_t \leftarrow \sigma_t \epsilon + (1 - \sigma_t) stopgrad(\boldsymbol{\hat z})$

    $\mathcal{L}_{Lora} \leftarrow \mathcal{L}_{Diff}(\boldsymbol{\hat z}_t,c_y)$ // \cref{eq:LoRA}

    Update $\phi$ with $\mathcal{L}_{Lora}$
    
  }
\end{algorithm*}

{
    \small
    \bibliographystyle{ieeenat_fullname}
    \bibliography{main}

\begin{thebibliography}{68}
\providecommand{\natexlab}[1]{#1}
\providecommand{\url}[1]{\texttt{#1}}
\expandafter\ifx\csname urlstyle\endcsname\relax
  \providecommand{\doi}[1]{doi: #1}\else
  \providecommand{\doi}{doi: \begingroup \urlstyle{rm}\Url}\fi

\bibitem[Agustsson and Timofte(2017)]{agustsson2017ntire}
Eirikur Agustsson and Radu Timofte.
\newblock Ntire 2017 challenge on single image super-resolution: Dataset and study.
\newblock In \emph{Proceedings of the IEEE conference on computer vision and pattern recognition workshops}, pages 126--135, 2017.

\bibitem[Arjovsky et~al.(2017)Arjovsky, Chintala, and Bottou]{arjovsky2017wasserstein}
Martin Arjovsky, Soumith Chintala, and L{\'e}on Bottou.
\newblock Wasserstein generative adversarial networks.
\newblock In \emph{International conference on machine learning}, pages 214--223. PMLR, 2017.

\bibitem[Blau and Michaeli(2018)]{blau2018perception}
Yochai Blau and Tomer Michaeli.
\newblock The perception-distortion tradeoff.
\newblock In \emph{Proceedings of the IEEE conference on computer vision and pattern recognition}, pages 6228--6237, 2018.

\bibitem[Cai et~al.(2019)Cai, Zeng, Yong, Cao, and Zhang]{cai2019toward}
Jianrui Cai, Hui Zeng, Hongwei Yong, Zisheng Cao, and Lei Zhang.
\newblock Toward real-world single image super-resolution: A new benchmark and a new model.
\newblock In \emph{Proceedings of the IEEE/CVF international conference on computer vision}, pages 3086--3095, 2019.

\bibitem[Chen et~al.(2022)Chen, Shi, Qin, Li, Han, Yang, and Guo]{chen2022real}
Chaofeng Chen, Xinyu Shi, Yipeng Qin, Xiaoming Li, Xiaoguang Han, Tao Yang, and Shihui Guo.
\newblock Real-world blind super-resolution via feature matching with implicit high-resolution priors.
\newblock In \emph{Proceedings of the 30th ACM International Conference on Multimedia}, pages 1329--1338, 2022.

\bibitem[Chen et~al.(2021)Chen, Wang, Guo, Xu, Deng, Liu, Ma, Xu, Xu, and Gao]{chen2021pre}
Hanting Chen, Yunhe Wang, Tianyu Guo, Chang Xu, Yiping Deng, Zhenhua Liu, Siwei Ma, Chunjing Xu, Chao Xu, and Wen Gao.
\newblock Pre-trained image processing transformer.
\newblock In \emph{Proceedings of the IEEE/CVF conference on computer vision and pattern recognition}, pages 12299--12310, 2021.

\bibitem[Ding et~al.(2020)Ding, Ma, Wang, and Simoncelli]{ding2020image}
Keyan Ding, Kede Ma, Shiqi Wang, and Eero~P Simoncelli.
\newblock Image quality assessment: Unifying structure and texture similarity.
\newblock \emph{IEEE transactions on pattern analysis and machine intelligence}, 44\penalty0 (5):\penalty0 2567--2581, 2020.

\bibitem[Dong et~al.(2014)Dong, Loy, He, and Tang]{dong2014learning}
Chao Dong, Chen~Change Loy, Kaiming He, and Xiaoou Tang.
\newblock Learning a deep convolutional network for image super-resolution.
\newblock In \emph{Computer Vision--ECCV 2014: 13th European Conference, Zurich, Switzerland, September 6-12, 2014, Proceedings, Part IV 13}, pages 184--199. Springer, 2014.

\bibitem[Dong et~al.(2015)Dong, Loy, He, and Tang]{dong2015image}
Chao Dong, Chen~Change Loy, Kaiming He, and Xiaoou Tang.
\newblock Image super-resolution using deep convolutional networks.
\newblock \emph{IEEE transactions on pattern analysis and machine intelligence}, 38\penalty0 (2):\penalty0 295--307, 2015.

\bibitem[Esser et~al.(2024)Esser, Kulal, Blattmann, Entezari, M{\"u}ller, Saini, Levi, Lorenz, Sauer, Boesel, et~al.]{esser2024scaling}
Patrick Esser, Sumith Kulal, Andreas Blattmann, Rahim Entezari, Jonas M{\"u}ller, Harry Saini, Yam Levi, Dominik Lorenz, Axel Sauer, Frederic Boesel, et~al.
\newblock Scaling rectified flow transformers for high-resolution image synthesis.
\newblock In \emph{Forty-first International Conference on Machine Learning}, 2024.

\bibitem[Goodfellow et~al.(2014)Goodfellow, Pouget-Abadie, Mirza, Xu, Warde-Farley, Ozair, Courville, and Bengio]{goodfellow2014generative}
Ian Goodfellow, Jean Pouget-Abadie, Mehdi Mirza, Bing Xu, David Warde-Farley, Sherjil Ozair, Aaron Courville, and Yoshua Bengio.
\newblock Generative adversarial nets.
\newblock \emph{Advances in neural information processing systems}, 27, 2014.

\bibitem[Gou et~al.(2021)Gou, Yu, Maybank, and Tao]{gou2021knowledge}
Jianping Gou, Baosheng Yu, Stephen~J Maybank, and Dacheng Tao.
\newblock Knowledge distillation: A survey.
\newblock \emph{International Journal of Computer Vision}, 129\penalty0 (6):\penalty0 1789--1819, 2021.

\bibitem[He and Cheng(2022)]{he2022revisiting}
Xiangyu He and Jian Cheng.
\newblock Revisiting l1 loss in super-resolution: a probabilistic view and beyond.
\newblock \emph{arXiv preprint arXiv:2201.10084}, 2022.

\bibitem[Heusel et~al.(2017)Heusel, Ramsauer, Unterthiner, Nessler, and Hochreiter]{heusel2017gans}
Martin Heusel, Hubert Ramsauer, Thomas Unterthiner, Bernhard Nessler, and Sepp Hochreiter.
\newblock Gans trained by a two time-scale update rule converge to a local nash equilibrium.
\newblock \emph{Advances in neural information processing systems}, 30, 2017.

\bibitem[Hinton(2015)]{hinton2015distilling}
Geoffrey Hinton.
\newblock Distilling the knowledge in a neural network.
\newblock \emph{arXiv preprint arXiv:1503.02531}, 2015.

\bibitem[Ho and Salimans(2022)]{ho2022classifier}
Jonathan Ho and Tim Salimans.
\newblock Classifier-free diffusion guidance.
\newblock \emph{arXiv preprint arXiv:2207.12598}, 2022.

\bibitem[Ho et~al.(2020)Ho, Jain, and Abbeel]{ho2020denoising}
Jonathan Ho, Ajay Jain, and Pieter Abbeel.
\newblock Denoising diffusion probabilistic models.
\newblock \emph{Advances in neural information processing systems}, 33:\penalty0 6840--6851, 2020.

\bibitem[Howard(2017)]{howard2017mobilenets}
Andrew~G Howard.
\newblock Mobilenets: Efficient convolutional neural networks for mobile vision applications.
\newblock \emph{arXiv preprint arXiv:1704.04861}, 2017.

\bibitem[Hu et~al.(2021)Hu, Shen, Wallis, Allen-Zhu, Li, Wang, Wang, and Chen]{hu2021lora}
Edward~J Hu, Yelong Shen, Phillip Wallis, Zeyuan Allen-Zhu, Yuanzhi Li, Shean Wang, Lu Wang, and Weizhu Chen.
\newblock Lora: Low-rank adaptation of large language models.
\newblock \emph{arXiv preprint arXiv:2106.09685}, 2021.

\bibitem[Karras et~al.(2019)Karras, Laine, and Aila]{karras2019style}
Tero Karras, Samuli Laine, and Timo Aila.
\newblock A style-based generator architecture for generative adversarial networks.
\newblock In \emph{Proceedings of the IEEE/CVF conference on computer vision and pattern recognition}, pages 4401--4410, 2019.

\bibitem[Kawar et~al.(2022)Kawar, Elad, Ermon, and Song]{kawar2022denoising}
Bahjat Kawar, Michael Elad, Stefano Ermon, and Jiaming Song.
\newblock Denoising diffusion restoration models.
\newblock \emph{Advances in Neural Information Processing Systems}, 35:\penalty0 23593--23606, 2022.

\bibitem[Ke et~al.(2021)Ke, Wang, Wang, Milanfar, and Yang]{ke2021musiq}
Junjie Ke, Qifei Wang, Yilin Wang, Peyman Milanfar, and Feng Yang.
\newblock Musiq: Multi-scale image quality transformer.
\newblock In \emph{Proceedings of the IEEE/CVF international conference on computer vision}, pages 5148--5157, 2021.

\bibitem[Kim et~al.(2016)Kim, Lee, and Lee]{kim2016accurate}
Jiwon Kim, Jung~Kwon Lee, and Kyoung~Mu Lee.
\newblock Accurate image super-resolution using very deep convolutional networks.
\newblock In \emph{Proceedings of the IEEE conference on computer vision and pattern recognition}, pages 1646--1654, 2016.

\bibitem[Kingma(2013)]{kingma2013auto}
Diederik~P Kingma.
\newblock Auto-encoding variational bayes.
\newblock \emph{arXiv preprint arXiv:1312.6114}, 2013.

\bibitem[Kullback and Leibler(1951)]{kullback1951information}
Solomon Kullback and Richard~A Leibler.
\newblock On information and sufficiency.
\newblock \emph{The annals of mathematical statistics}, 22\penalty0 (1):\penalty0 79--86, 1951.

\bibitem[Ledig et~al.(2017)Ledig, Theis, Husz{\'a}r, Caballero, Cunningham, Acosta, Aitken, Tejani, Totz, Wang, et~al.]{ledig2017photo}
Christian Ledig, Lucas Theis, Ferenc Husz{\'a}r, Jose Caballero, Andrew Cunningham, Alejandro Acosta, Andrew Aitken, Alykhan Tejani, Johannes Totz, Zehan Wang, et~al.
\newblock Photo-realistic single image super-resolution using a generative adversarial network.
\newblock In \emph{Proceedings of the IEEE conference on computer vision and pattern recognition}, pages 4681--4690, 2017.

\bibitem[Li et~al.(2023)Li, Zhang, Liang, Cao, Liu, Gong, Zhang, Tang, Liu, Demandolx, et~al.]{li2023lsdir}
Yawei Li, Kai Zhang, Jingyun Liang, Jiezhang Cao, Ce Liu, Rui Gong, Yulun Zhang, Hao Tang, Yun Liu, Denis Demandolx, et~al.
\newblock Lsdir: A large scale dataset for image restoration.
\newblock In \emph{Proceedings of the IEEE/CVF Conference on Computer Vision and Pattern Recognition}, pages 1775--1787, 2023.

\bibitem[Liang et~al.(2022)Liang, Zeng, and Zhang]{liang2022details}
Jie Liang, Hui Zeng, and Lei Zhang.
\newblock Details or artifacts: A locally discriminative learning approach to realistic image super-resolution.
\newblock In \emph{Proceedings of the IEEE/CVF Conference on Computer Vision and Pattern Recognition}, pages 5657--5666, 2022.

\bibitem[Lin et~al.(2023)Lin, He, Chen, Lyu, Dai, Yu, Ouyang, Qiao, and Dong]{lin2023diffbir}
Xinqi Lin, Jingwen He, Ziyan Chen, Zhaoyang Lyu, Bo Dai, Fanghua Yu, Wanli Ouyang, Yu Qiao, and Chao Dong.
\newblock Diffbir: Towards blind image restoration with generative diffusion prior.
\newblock \emph{arXiv preprint arXiv:2308.15070}, 2023.

\bibitem[Loshchilov(2017)]{loshchilov2017decoupled}
I Loshchilov.
\newblock Decoupled weight decay regularization.
\newblock \emph{arXiv preprint arXiv:1711.05101}, 2017.

\bibitem[Luo et~al.(2023)Luo, Tan, Huang, Li, and Zhao]{luo2023latent}
Simian Luo, Yiqin Tan, Longbo Huang, Jian Li, and Hang Zhao.
\newblock Latent consistency models: Synthesizing high-resolution images with few-step inference.
\newblock \emph{arXiv preprint arXiv:2310.04378}, 2023.

\bibitem[Luo et~al.(2024)Luo, Xie, Qu, and Fu]{luo2024skipdiff}
Xiaotong Luo, Yuan Xie, Yanyun Qu, and Yun Fu.
\newblock Skipdiff: Adaptive skip diffusion model for high-fidelity perceptual image super-resolution.
\newblock In \emph{Proceedings of the AAAI Conference on Artificial Intelligence}, pages 4017--4025, 2024.

\bibitem[Mirza(2014)]{mirza2014conditional}
Mehdi Mirza.
\newblock Conditional generative adversarial nets.
\newblock \emph{arXiv preprint arXiv:1411.1784}, 2014.

\bibitem[Nguyen and Tran(2024)]{nguyen2024swiftbrush}
Thuan~Hoang Nguyen and Anh Tran.
\newblock Swiftbrush: One-step text-to-image diffusion model with variational score distillation.
\newblock In \emph{Proceedings of the IEEE/CVF Conference on Computer Vision and Pattern Recognition}, pages 7807--7816, 2024.

\bibitem[Podell et~al.(2023)Podell, English, Lacey, Blattmann, Dockhorn, M{\"u}ller, Penna, and Rombach]{podell2023sdxl}
Dustin Podell, Zion English, Kyle Lacey, Andreas Blattmann, Tim Dockhorn, Jonas M{\"u}ller, Joe Penna, and Robin Rombach.
\newblock Sdxl: Improving latent diffusion models for high-resolution image synthesis.
\newblock \emph{arXiv preprint arXiv:2307.01952}, 2023.

\bibitem[Poole et~al.(2022)Poole, Jain, Barron, and Mildenhall]{poole2022dreamfusion}
Ben Poole, Ajay Jain, Jonathan~T Barron, and Ben Mildenhall.
\newblock Dreamfusion: Text-to-3d using 2d diffusion.
\newblock \emph{arXiv preprint arXiv:2209.14988}, 2022.

\bibitem[Radford(2015)]{radford2015unsupervised}
Alec Radford.
\newblock Unsupervised representation learning with deep convolutional generative adversarial networks.
\newblock \emph{arXiv preprint arXiv:1511.06434}, 2015.

\bibitem[Ramesh et~al.(2022)Ramesh, Dhariwal, Nichol, Chu, and Chen]{ramesh2022hierarchical}
Aditya Ramesh, Prafulla Dhariwal, Alex Nichol, Casey Chu, and Mark Chen.
\newblock Hierarchical text-conditional image generation with clip latents.
\newblock \emph{arXiv preprint arXiv:2204.06125}, 1\penalty0 (2):\penalty0 3, 2022.

\bibitem[Roeder et~al.(2017)Roeder, Wu, and Duvenaud]{roeder2017sticking}
Geoffrey Roeder, Yuhuai Wu, and David~K Duvenaud.
\newblock Sticking the landing: Simple, lower-variance gradient estimators for variational inference.
\newblock \emph{Advances in Neural Information Processing Systems}, 30, 2017.

\bibitem[Rombach et~al.(2022)Rombach, Blattmann, Lorenz, Esser, and Ommer]{rombach2022high}
Robin Rombach, Andreas Blattmann, Dominik Lorenz, Patrick Esser, and Bj{\"o}rn Ommer.
\newblock High-resolution image synthesis with latent diffusion models.
\newblock In \emph{Proceedings of the IEEE/CVF conference on computer vision and pattern recognition}, pages 10684--10695, 2022.

\bibitem[Sauer et~al.(2023)Sauer, Lorenz, Blattmann, and Rombach]{sauer2023adversarial}
Axel Sauer, Dominik Lorenz, Andreas Blattmann, and Robin Rombach.
\newblock Adversarial diffusion distillation.
\newblock \emph{arXiv preprint arXiv:2311.17042}, 2023.

\bibitem[Song et~al.(2020)Song, Sohl-Dickstein, Kingma, Kumar, Ermon, and Poole]{song2020score}
Yang Song, Jascha Sohl-Dickstein, Diederik~P Kingma, Abhishek Kumar, Stefano Ermon, and Ben Poole.
\newblock Score-based generative modeling through stochastic differential equations.
\newblock \emph{arXiv preprint arXiv:2011.13456}, 2020.

\bibitem[Timofte et~al.(2017)Timofte, Agustsson, Van~Gool, Yang, and Zhang]{timofte2017ntire}
Radu Timofte, Eirikur Agustsson, Luc Van~Gool, Ming-Hsuan Yang, and Lei Zhang.
\newblock Ntire 2017 challenge on single image super-resolution: Methods and results.
\newblock In \emph{Proceedings of the IEEE conference on computer vision and pattern recognition workshops}, pages 114--125, 2017.

\bibitem[Wang et~al.(2023)Wang, Chan, and Loy]{wang2023exploring}
Jianyi Wang, Kelvin~CK Chan, and Chen~Change Loy.
\newblock Exploring clip for assessing the look and feel of images.
\newblock In \emph{Proceedings of the AAAI Conference on Artificial Intelligence}, pages 2555--2563, 2023.

\bibitem[Wang et~al.(2024{\natexlab{a}})Wang, Yue, Zhou, Chan, and Loy]{wang2024exploiting}
Jianyi Wang, Zongsheng Yue, Shangchen Zhou, Kelvin~CK Chan, and Chen~Change Loy.
\newblock Exploiting diffusion prior for real-world image super-resolution.
\newblock \emph{International Journal of Computer Vision}, pages 1--21, 2024{\natexlab{a}}.

\bibitem[Wang et~al.(2018)Wang, Yu, Wu, Gu, Liu, Dong, Qiao, and Change~Loy]{wang2018esrgan}
Xintao Wang, Ke Yu, Shixiang Wu, Jinjin Gu, Yihao Liu, Chao Dong, Yu Qiao, and Chen Change~Loy.
\newblock Esrgan: Enhanced super-resolution generative adversarial networks.
\newblock In \emph{Proceedings of the European conference on computer vision (ECCV) workshops}, pages 0--0, 2018.

\bibitem[Wang et~al.(2021)Wang, Xie, Dong, and Shan]{wang2021real}
Xintao Wang, Liangbin Xie, Chao Dong, and Ying Shan.
\newblock Real-esrgan: Training real-world blind super-resolution with pure synthetic data.
\newblock In \emph{Proceedings of the IEEE/CVF international conference on computer vision}, pages 1905--1914, 2021.

\bibitem[Wang et~al.(2022)Wang, Yu, and Zhang]{wang2022zero}
Yinhuai Wang, Jiwen Yu, and Jian Zhang.
\newblock Zero-shot image restoration using denoising diffusion null-space model.
\newblock \emph{arXiv preprint arXiv:2212.00490}, 2022.

\bibitem[Wang et~al.(2024{\natexlab{b}})Wang, Yang, Chen, Wang, Guo, Chau, Liu, Qiao, Kot, and Wen]{wang2024sinsr}
Yufei Wang, Wenhan Yang, Xinyuan Chen, Yaohui Wang, Lanqing Guo, Lap-Pui Chau, Ziwei Liu, Yu Qiao, Alex~C Kot, and Bihan Wen.
\newblock Sinsr: diffusion-based image super-resolution in a single step.
\newblock In \emph{Proceedings of the IEEE/CVF Conference on Computer Vision and Pattern Recognition}, pages 25796--25805, 2024{\natexlab{b}}.

\bibitem[Wang et~al.(2004)Wang, Bovik, Sheikh, and Simoncelli]{wang2004image}
Zhou Wang, Alan~C Bovik, Hamid~R Sheikh, and Eero~P Simoncelli.
\newblock Image quality assessment: from error visibility to structural similarity.
\newblock \emph{IEEE transactions on image processing}, 13\penalty0 (4):\penalty0 600--612, 2004.

\bibitem[Wang et~al.(2024{\natexlab{c}})Wang, Lu, Wang, Bao, Li, Su, and Zhu]{wang2024prolificdreamer}
Zhengyi Wang, Cheng Lu, Yikai Wang, Fan Bao, Chongxuan Li, Hang Su, and Jun Zhu.
\newblock Prolificdreamer: High-fidelity and diverse text-to-3d generation with variational score distillation.
\newblock \emph{Advances in Neural Information Processing Systems}, 36, 2024{\natexlab{c}}.

\bibitem[Wei et~al.(2020)Wei, Xie, Lu, Zhan, Ye, Zuo, and Lin]{wei2020component}
Pengxu Wei, Ziwei Xie, Hannan Lu, Zongyuan Zhan, Qixiang Ye, Wangmeng Zuo, and Liang Lin.
\newblock Component divide-and-conquer for real-world image super-resolution.
\newblock In \emph{Computer Vision--ECCV 2020: 16th European Conference, Glasgow, UK, August 23--28, 2020, Proceedings, Part VIII 16}, pages 101--117. Springer, 2020.

\bibitem[Wu et~al.(2024{\natexlab{a}})Wu, Sun, Ma, and Zhang]{wu2024one}
Rongyuan Wu, Lingchen Sun, Zhiyuan Ma, and Lei Zhang.
\newblock One-step effective diffusion network for real-world image super-resolution.
\newblock \emph{arXiv preprint arXiv:2406.08177}, 2024{\natexlab{a}}.

\bibitem[Wu et~al.(2024{\natexlab{b}})Wu, Yang, Sun, Zhang, Li, and Zhang]{wu2024seesr}
Rongyuan Wu, Tao Yang, Lingchen Sun, Zhengqiang Zhang, Shuai Li, and Lei Zhang.
\newblock Seesr: Towards semantics-aware real-world image super-resolution.
\newblock In \emph{Proceedings of the IEEE/CVF conference on computer vision and pattern recognition}, pages 25456--25467, 2024{\natexlab{b}}.

\bibitem[Xie et~al.(2024)Xie, Tai, Zhang, Zhang, Zhou, and Yang]{xie2024addsr}
Rui Xie, Ying Tai, Kai Zhang, Zhenyu Zhang, Jun Zhou, and Jian Yang.
\newblock Addsr: Accelerating diffusion-based blind super-resolution with adversarial diffusion distillation.
\newblock \emph{arXiv preprint arXiv:2404.01717}, 2024.

\bibitem[Yang et~al.(2022)Yang, Wu, Shi, Lao, Gong, Cao, Wang, and Yang]{yang2022maniqa}
Sidi Yang, Tianhe Wu, Shuwei Shi, Shanshan Lao, Yuan Gong, Mingdeng Cao, Jiahao Wang, and Yujiu Yang.
\newblock Maniqa: Multi-dimension attention network for no-reference image quality assessment.
\newblock In \emph{Proceedings of the IEEE/CVF Conference on Computer Vision and Pattern Recognition}, pages 1191--1200, 2022.

\bibitem[Yang et~al.(2023)Yang, Wu, Ren, Xie, and Zhang]{yang2023pixel}
Tao Yang, Rongyuan Wu, Peiran Ren, Xuansong Xie, and Lei Zhang.
\newblock Pixel-aware stable diffusion for realistic image super-resolution and personalized stylization.
\newblock \emph{arXiv preprint arXiv:2308.14469}, 2023.

\bibitem[Yim et~al.(2017)Yim, Joo, Bae, and Kim]{yim2017gift}
Junho Yim, Donggyu Joo, Jihoon Bae, and Junmo Kim.
\newblock A gift from knowledge distillation: Fast optimization, network minimization and transfer learning.
\newblock In \emph{Proceedings of the IEEE conference on computer vision and pattern recognition}, pages 4133--4141, 2017.

\bibitem[Yin et~al.(2024{\natexlab{a}})Yin, Gharbi, Park, Zhang, Shechtman, Durand, and Freeman]{yin2024improved}
Tianwei Yin, Micha{\"e}l Gharbi, Taesung Park, Richard Zhang, Eli Shechtman, Fredo Durand, and William~T Freeman.
\newblock Improved distribution matching distillation for fast image synthesis.
\newblock \emph{arXiv preprint arXiv:2405.14867}, 2024{\natexlab{a}}.

\bibitem[Yin et~al.(2024{\natexlab{b}})Yin, Gharbi, Zhang, Shechtman, Durand, Freeman, and Park]{yin2024one}
Tianwei Yin, Micha{\"e}l Gharbi, Richard Zhang, Eli Shechtman, Fredo Durand, William~T Freeman, and Taesung Park.
\newblock One-step diffusion with distribution matching distillation.
\newblock In \emph{Proceedings of the IEEE/CVF Conference on Computer Vision and Pattern Recognition}, pages 6613--6623, 2024{\natexlab{b}}.

\bibitem[Yu et~al.(2024)Yu, Gu, Li, Hu, Kong, Wang, He, Qiao, and Dong]{yu2024scaling}
Fanghua Yu, Jinjin Gu, Zheyuan Li, Jinfan Hu, Xiangtao Kong, Xintao Wang, Jingwen He, Yu Qiao, and Chao Dong.
\newblock Scaling up to excellence: Practicing model scaling for photo-realistic image restoration in the wild.
\newblock In \emph{Proceedings of the IEEE/CVF Conference on Computer Vision and Pattern Recognition}, pages 25669--25680, 2024.

\bibitem[Yue et~al.(2024)Yue, Wang, and Loy]{yue2024resshift}
Zongsheng Yue, Jianyi Wang, and Chen~Change Loy.
\newblock Resshift: Efficient diffusion model for image super-resolution by residual shifting.
\newblock \emph{Advances in Neural Information Processing Systems}, 36, 2024.

\bibitem[Zhang et~al.(2021)Zhang, Liang, Van~Gool, and Timofte]{zhang2021designing}
Kai Zhang, Jingyun Liang, Luc Van~Gool, and Radu Timofte.
\newblock Designing a practical degradation model for deep blind image super-resolution.
\newblock In \emph{Proceedings of the IEEE/CVF International Conference on Computer Vision}, pages 4791--4800, 2021.

\bibitem[Zhang et~al.(2015)Zhang, Zhang, and Bovik]{zhang2015feature}
Lin Zhang, Lei Zhang, and Alan~C Bovik.
\newblock A feature-enriched completely blind image quality evaluator.
\newblock \emph{IEEE Transactions on Image Processing}, 24\penalty0 (8):\penalty0 2579--2591, 2015.

\bibitem[Zhang et~al.(2023)Zhang, Rao, and Agrawala]{zhang2023adding}
Lvmin Zhang, Anyi Rao, and Maneesh Agrawala.
\newblock Adding conditional control to text-to-image diffusion models.
\newblock In \emph{Proceedings of the IEEE/CVF International Conference on Computer Vision}, pages 3836--3847, 2023.

\bibitem[Zhang et~al.(2018)Zhang, Isola, Efros, Shechtman, and Wang]{zhang2018unreasonable}
Richard Zhang, Phillip Isola, Alexei~A Efros, Eli Shechtman, and Oliver Wang.
\newblock The unreasonable effectiveness of deep features as a perceptual metric.
\newblock In \emph{Proceedings of the IEEE conference on computer vision and pattern recognition}, pages 586--595, 2018.

\bibitem[Zhang et~al.(2022{\natexlab{a}})Zhang, Zeng, Guo, and Zhang]{zhang2022efficient}
Xindong Zhang, Hui Zeng, Shi Guo, and Lei Zhang.
\newblock Efficient long-range attention network for image super-resolution.
\newblock In \emph{European conference on computer vision}, pages 649--667. Springer, 2022{\natexlab{a}}.

\bibitem[Zhang et~al.(2022{\natexlab{b}})Zhang, Ji, Hao, and Yao]{zhang2022perception}
Yuehan Zhang, Bo Ji, Jia Hao, and Angela Yao.
\newblock Perception-distortion balanced admm optimization for single-image super-resolution.
\newblock In \emph{European Conference on Computer Vision}, pages 108--125. Springer, 2022{\natexlab{b}}.

\end{thebibliography}
}

\end{document}